\documentclass[11pt]{article}

\usepackage[final]{acl}

\usepackage{times}
\usepackage{latexsym}
\usepackage{amsfonts}
\usepackage{subcaption}
\usepackage{booktabs}
\usepackage{multirow}
\usepackage{amsmath}

\usepackage[T1]{fontenc}

\usepackage[utf8]{inputenc}

\usepackage{microtype}

\usepackage{inconsolata}

\usepackage{graphicx}

\title{SV-Detect: AI-generated Text Detection with Steering Vectors}

\author{Mikhail Vishnyakov \\
  Independent Researcher \\
  \texttt{mikhail.n.vishnyakov@gmail.com} \\\And
  Tatiana Gaintseva \\
  Queen Mary University of London \\
  \texttt{t.gaintseva@qmul.ac.uk} \\}

\begin{document}
\maketitle

\begin{abstract}
Detecting AI-generated text is especially difficult under distribution shift, such as transfer across domains, source models, and editing attacks. We propose an AI-generated text detector based on \emph{steering vectors} extracted from the hidden representations of a frozen language model. At each layer, we construct a direction that separates human-written from AI-generated text, and represent each input by its layer-wise alignment with these directions. A lightweight classifier trained on these projection features yields the final detection score. Our method achieves strong performance both in-distribution and under distribution shift, including across domains, source models, and machine-editing transformations such as polishing and rewriting. 
Interpretation analyses show that the learned directions align with recognizable stylistic cues while capturing substantial additional signal beyond surface features. These results position AI-generated text detection as a representation-space probing problem and show that steering vectors provide a simple and effective solution. 
Code is available at \href{https://github.com/Atmyre/sv-detect/}{https://github.com/Atmyre/sv-detect/}
\end{abstract}

\section{Introduction}


The rapid deployment of large language models has made AI-generated text increasingly fluent, diverse, and difficult to distinguish from human writing.~\citep{DBLP:journals/corr/abs-2410-23746,WuYZYCW25}. This creates practical challenges for content moderation, authorship verification, and benchmark integrity, and has led to growing interest in methods for AI-generated text detection~\citep{WuYZYCW25,DBLP:journals/corr/abs-2306-15666}. At the same time, the problem has become harder: modern detectors must operate not only on direct generations, but also under distribution shift, including transfer across domains, source models, and editing-based attacks such as paraphrasing, polishing, or rewriting \citep{DBLP:journals/csr/KehkashanRAAAHA25}.
\smallskip

\begin{figure}[t]
      \centering
      \includegraphics[width=\columnwidth]{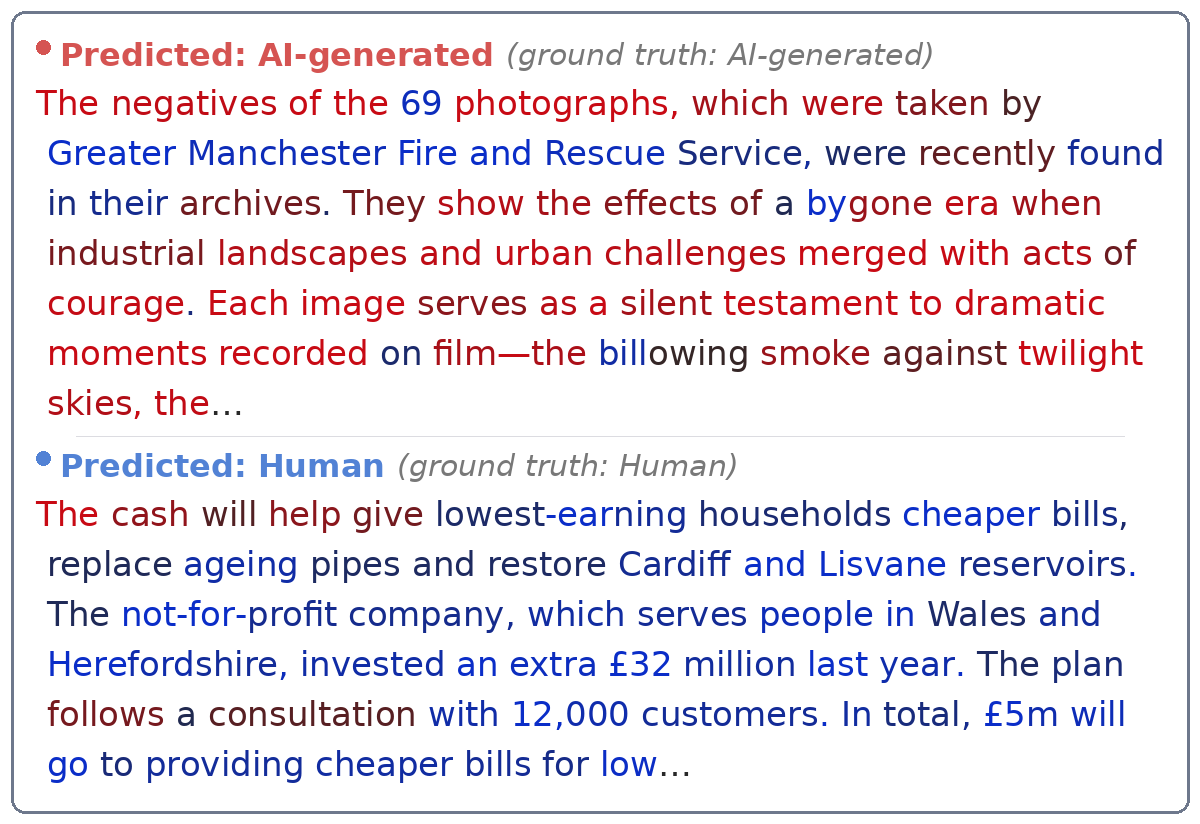}
      \caption{Token-level steering vector projections distinguish LLM-generated
      text (top) from human writing (bottom). Word saturation reflects each
      token's signed contribution to the classifier.}
      \label{fig:teaser}
      \vspace{-0.5cm}
  \end{figure}

A large body of prior work approaches this problem through token-level statistics, likelihood-based scores, perturbation tests~\citep{DBLP:journals/corr/abs-2310-05130,DBLP:journals/corr/abs-2305-17359,DBLP:journals/corr/abs-2401-12070}, or supervised classifiers trained directly on text~\citep{DBLP:conf/coling/JawaharASL20,DBLP:journals/corr/abs-2412-10432}. While these methods can be effective in-distribution, they often degrade under transfer, suggesting that many detectors rely on artifacts tied to a particular generator, dataset, or attack style~\citep{DBLP:conf/acl/LiLCBWWY0024,DBLP:journals/corr/abs-2410-23746,padben}. This motivates a different perspective: instead of using solely surface form or output probabilities, can we identify a more stable signal in the \emph{intermediate representations} of a frozen LLM?

In this paper, we propose \textbf{SV-Detect}, an AI-generated text detector based on \emph{steering vectors} — AI-generated text directions extracted from the hidden states of a frozen language model. Our central hypothesis is that \textbf{human-written and AI-generated texts induce systematically different directions in representation space}, and these directions can be used as robust detection features. Concretely, for each transformer layer, we construct a vector that separates human and AI-generated texts, represent a new input by its layer-wise alignment with these directions, and train a lightweight classifier on the resulting projection features. This yields a simple detector that does not require fine-tuning the underlying language model and naturally supports interpretation 
at the level of layers and directions.
\smallskip

We evaluate SV-Detect on several complementary benchmarks. Our main evaluation covers \emph{DetectRL} \citep{DBLP:journals/corr/abs-2410-23746} and \emph{MIRAGE} \citep{DBLP:journals/corr/abs-2509-14268}: DetectRL tests robustness across domains, source models, and attack families, and MIRAGE targets direct generation and machine-assisted editing scenarios, including polishing and rewriting. 
%
%
We further assess out-of-distribution generalization by transferring detectors trained on \emph{DetectRL} and \emph{MIRAGE} across benchmarks, including transfer between the two and transfer to \emph{PADBen}~\citep{padben}.
Across these settings, SV-Detect achieves near-perfect in-distribution performance, robust transfer, and strong cross-benchmark generalization. It also substantially outperforms RepreGuard~\citep{RepreGuard}, the closest representation-based detector, under matched-backbone comparisons.

\smallskip

Beyond accuracy, we analyze what SV-Detect is actually using. The learned steering directions align with interpretable lexical and stylistic cues, but they also capture substantial signal beyond hand-picked surface features. This suggests that SV-Detect is not only effective, but also a useful tool for studying how machine-generated text differs from human writing at the representation level.
\smallskip
\vspace{-0.5cm}

\noindent Overall, our contributions are as follows:
\begin{itemize}
    \item We \textbf{introduce SV-Detect}, a simple and effective AI-generated text detector based on directions extracted from the hidden representations of a frozen language model.
    \vspace{-0.2cm}
    \item We \textbf{demonstrate strong generalization} across a set of benchmarks, including cross-domain, cross-source, cross-attack and cross-benchmark transfer.
    \vspace{-0.2cm}
    \item We \textbf{provide interpretation analyses} showing that the learned directions align with meaningful lexical and stylistic cues while also encoding additional representation-level signal.
\end{itemize}

\noindent Overall, our results suggest that AI-generated text detection can be viewed as a \emph{representation-space probing problem}, where machine-generated text is identified by the directions along which it differs from human writing in hidden-state space.
\section{Related Work}

\noindent \textbf{Supervised detection of AI-generated text.}
A common approach to AI-generated text detection is to train a classifier on human-written and model-generated corpora. These methods typically use pretrained encoders such as BERT and often achieve strong in-domain performance when train and test distributions are matched~\citep{DBLP:conf/coling/JawaharASL20,DBLP:journals/corr/abs-2412-10432}. Variants include boundary-based and topology-aware detectors~\citep{kushnareva2024aigeneratedtextboundarydetection,DBLP:journals/corr/abs-2109-04825}, and methods that improve transfer by removing brittle components of encoder representations~\citep{DBLP:journals/corr/abs-2410-08113}. 
A closely related recent method, \emph{RepreGuard}~\citep{RepreGuard}, also uses hidden-state features from a frozen LM rather than fine-tuning an encoder. Unlike RepreGuard, we build \emph{layer-wise discriminative directions} and represent each text by its alignment with these directions at every layer, which yields a compact layer-diagnostic representation and supports interpretability analyses.
%
\smallskip

\vspace{-0.5cm}
\noindent \textbf{Zero-shot and score-based detectors.}
Another line of work aims to detect AI-generated text without training a dedicated classifier. Early methods rely on token-level statistics such as likelihood, entropy, rank, log-rank, and likelihood-ratio scores~\citep{DBLP:journals/corr/abs-2306-05540,DBLP:journals/corr/abs-2410-23746}. More recent methods derive zero-shot criteria from a reference LM, including DetectGPT~\citep{DBLP:journals/corr/abs-2301-11305}, Fast-DetectGPT~\citep{DBLP:journals/corr/abs-2310-05130}, DNA-GPT~\citep{DBLP:journals/corr/abs-2305-17359}, and Binoculars~\citep{DBLP:journals/corr/abs-2401-12070}. In contrast, SV-Detect is not zero-shot: it learns a lightweight detector from representation-level features rather than directly from text.
\smallskip


\noindent \textbf{Robustness and evaluation under distribution shift.}
Recent work has emphasized that strong in-domain results are not sufficient for realistic machine-generated text detection~\citep{DBLP:conf/naacl/TuftsZL25,DBLP:conf/acl/LiLCBWWY0024,kushnareva2024aigeneratedtextboundarydetection}. \emph{DetectRL}~\citep{DBLP:journals/corr/abs-2410-23746} benchmarks transfer across domains, source LLMs, and attack families, while \emph{MIRAGE}~\citep{DBLP:journals/corr/abs-2509-14268} focuses on generation, polishing, and rewriting.
\emph{PADBen}~\citep{padben} further targets paraphrase-based laundering, iterative paraphrase depth, and human-edited AI-generated text, complementing the attack families covered by DetectRL and MIRAGE with a laundering-centric perspective.
Following this robustness-centered evaluation paradigm, we evaluate SV-Detect on these benchmarks and additionally test cross-benchmark transfer.
\section{Methodology}
\label{sec:methodology}

\begin{figure}[t]
      \centering
      \includegraphics[width=\columnwidth]{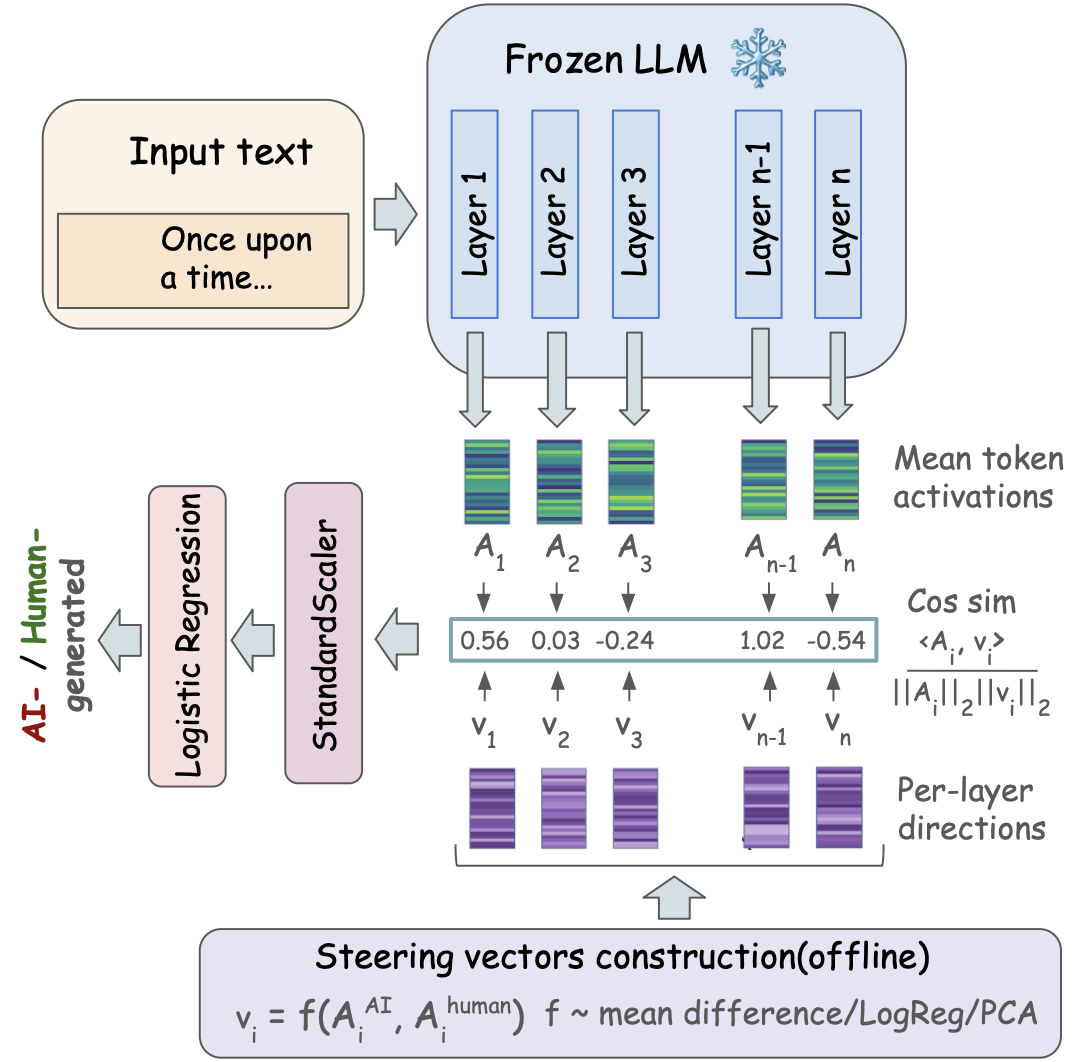}
      \caption{Overview of SV-Detect. A frozen LLM is used to extract mean-pooled hidden activations from each layer. Activations are projected onto layer-wise steering vectors, and the resulting cosine similarity scores are standardized and passed to a logistic regression classifier for AI-generated text detection.}
      \label{fig:method}
      \vspace{-0.5cm}
  \end{figure}

\subsection{Overview}

The key idea of our method is based on the hypothesis that human-written and AI-generated texts induce systematically different activation patterns inside a pretrained model. We capture these differences as directions in representation space and use them to construct features for a downstream detection model.
\smallskip

Our pipeline is illustrated in Fig.~\ref{fig:method} and consists of four stages:
(i) extracting layer-wise activations from a frozen language model,
(ii) extracting \emph{AI-generated text directions} that distinguish human-written and machine-generated texts,
(iii) projecting text representations onto these directions to obtain low-dimensional features, and
(iv) training a lightweight classifier on top of these features.

This design transforms high-dimensional hidden states into interpretable scores that quantify how strongly a text aligns with AI-generated text directions across the network.

\subsection{Layer-wise text representations}

Let $f$ be a frozen transformer language model with $L$ layers. Given an input text $x$ tokenized into $T$ tokens, let $H_l(x) \in \mathbb{R}^{T \times d}$
denote the hidden states at layer $l \in \{1,\dots,L\}$, where $d$ is the hidden dimension. We construct the text representation at layer $l$ by mean-pooling hidden states across the token dimension:
\[
a_l(x) = \frac{1}{T} \sum_{t=1}^{T} H_l(x)_t \in \mathbb{R}^d.
\]
The full representation of $x$ is therefore the collection of its representations at each layer:
\[
a(x) = \bigl(a_1(x), \dots, a_L(x)\bigr).
\]


\subsection{Constructing AI-generated text directions}

Suppose we are given training data that consists of a set of human-written texts $\mathcal{D}_{\mathrm{real}}$ and a set of AI-generated texts $\mathcal{D}_{\mathrm{fake}}$
\[
\mathcal{D}_{\mathrm{real}} = \{x_i^{(r)}\}_{i=1}^{N_r}, \mathcal{D}_{\mathrm{fake}} = \{x_j^{(f)}\}_{j=1}^{N_f}
\]

For each layer \(l\), we use the pooled activations
\[
a_l^{(r)} = \{a_l(x_i^{(r)})\}_{i=1}^{N_r},
\qquad
a_l^{(f)} = \{a_l(x_j^{(f)})\}_{j=1}^{N_f}
\]
to construct a vector \(v_l \in \mathbb{R}^d\) that captures the direction separating
AI-generated from human-written text representations.
We refer to these directions as \emph{steering vectors}: they are representation-space directions of the same kind used in activation steering and representation engineering work~\citep{repr_engineering}. However, note that in SV-Detect they are used only as probing and classification features.
%
We study three methods for constructing steering vectors, following common practice in representation engineering~\citep{repr_engineering}.

\paragraph{Mean-difference.}
The simplest choice is the normalized difference between class means:
\[
\mu_l^{(f)} = \frac{1}{N_f}\sum_{j=1}^{N_f} a_l(x_j^{(f)}),
\ 
\mu_l^{(r)} = \frac{1}{N_r}\sum_{i=1}^{N_r} a_l(x_i^{(r)}),
\]
\[
v_l^{\mathrm{mean}} =
\frac{\mu_l^{(f)} - \mu_l^{(r)}}
{\|\mu_l^{(f)} - \mu_l^{(r)}\|_2}.
\]
This vector points from the average human-written representation toward the average AI-generated representation. 

\paragraph{Logistic-regression.}
One alternative is to fit a linear classifier directly in the activation space of each layer and use the normal vector of the separating hyperplane. For each layer $l$, we train a logistic regression model to separate AI-generated and human-written text activations. Let $w_l$ denote the learned weight vector for layer $l$. We define the steering vector for this layer as its normalized version:
\[
v_l^{\mathrm{logreg}} = \frac{w_l}{\|w_l\|_2}.
\]

\paragraph{PCA.}
As a third variant, we compute paired differences between AI-generated and human-written text activations and define the steering vector as the leading principal component of these differences:
\[
v_l^{\mathrm{pca}} = \mathrm{PC}_1\Bigl(\{a_l(x_j^{(f)}) - a_l(x_i^{(r)})\}\Bigr).
\]
This construction aims to identify the dominant axis associated with the human-to-AI-generated shift in representation space.

\subsection{Projection features}

Once steering vectors are obtained, we construct a feature representation of a text by using its alignment with these directions across layers. For each layer, we compute the layer-wise score as a cosine similarity between the pooled activation of this layer and its steering vector
\[
s_l(x) = \frac{\langle a_l(x), v_l \rangle}{\|a_l(x)\|_2}.
\]

The final feature vector is obtained by concatenating scores from all layers:
\[
s(x) = \bigl(s_1(x), \dots, s_L(x)\bigr) \in \mathbb{R}^L.
\]
Thus, instead of classifying directly from the full hidden states, we classify from a compact representation that summarizes how strongly the text aligns with AI-generated text directions throughout the model.






\subsection{Detection head}

To actually decide if the text is AI-generated, its projection features are fed into a lightweight downstream classifier. Our default detector consists of feature standardization followed by logistic regression. Given a feature vector $s(x)$, the detector outputs
\[
p(y=1 \mid x) = \sigma\bigl(w^\top \tilde{s}(x) + b\bigr),
\]
where $\tilde{s}(x)$ is the standardized version of $s(x)$, $\sigma(\cdot)$ is the sigmoid function, and $y=1$ denotes machine-generated text.

This choice isolates the effect of the steering-based representation from the complexity of the classifier. 
We ablate our choice of classifier considering alternative models in Sec.~\ref{sec:ablation}.
\smallskip

At test time, we extract activations for the new text, compute its projection features, and apply the trained classifier. Importantly, the reference LLM remains frozen throughout the entire pipeline.
\section{Experiments}
\label{sec:experiments}

\begin{figure*}[t]
    \centering

    \begin{minipage}[t]{0.68\textwidth}
        \vspace{0pt}
        \centering

        \begin{subfigure}[t]{\linewidth}
            \centering
            \includegraphics[width=\linewidth]{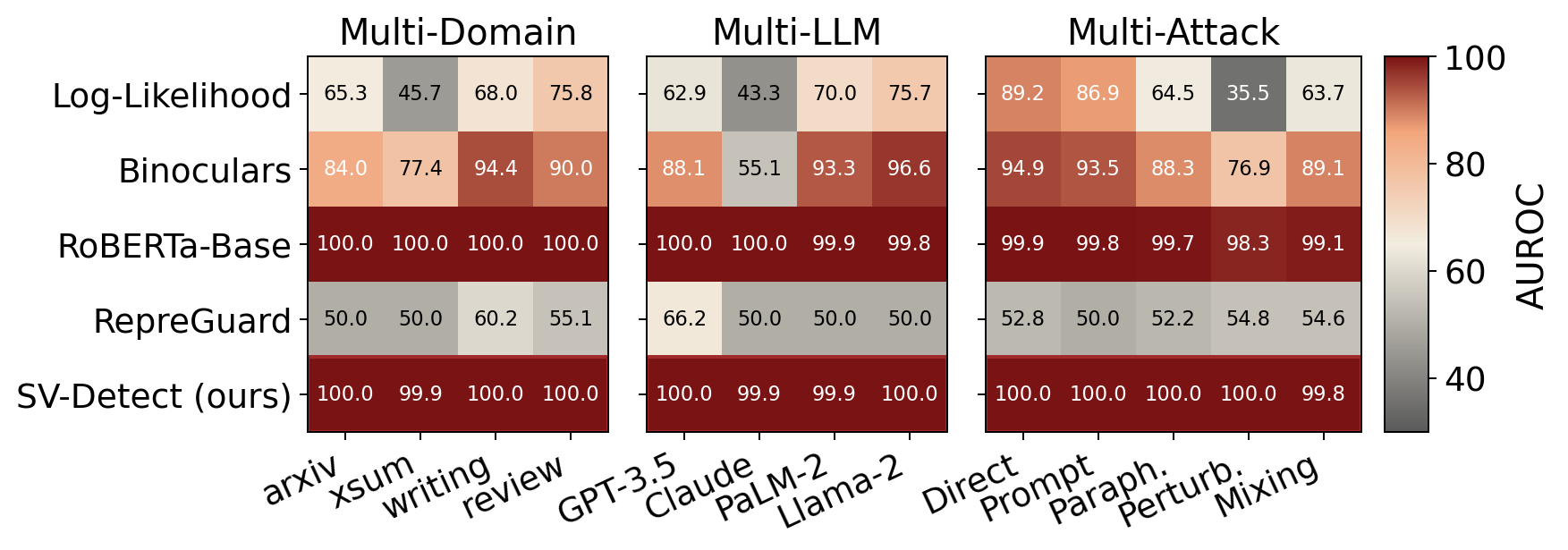}
            \caption{In-domain AUROC across the Multi-Domain, Multi-LLM, and Multi-Attack settings.}
            \label{fig:detectrl_indomain_auroc}
        \end{subfigure}

        \vspace{0.8em}

        \begin{subfigure}[t]{\linewidth}
            \centering
            \includegraphics[width=\linewidth]{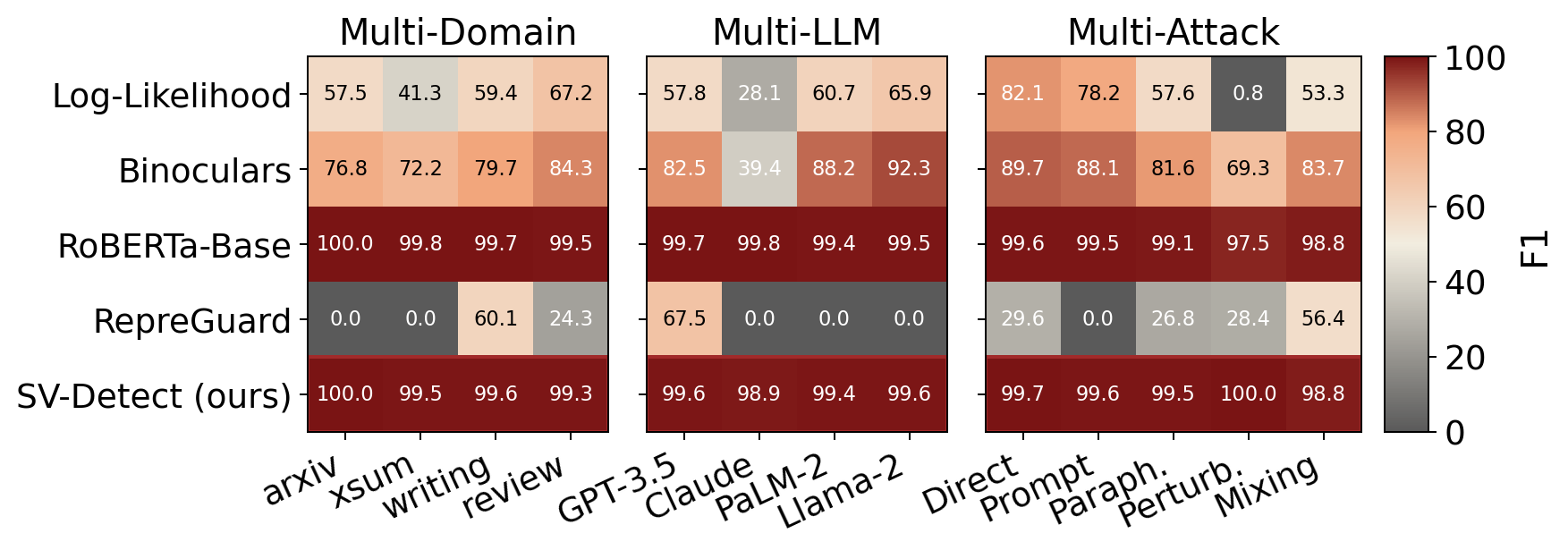}
            \caption{In-domain $F_1$ across the Multi-Domain, Multi-LLM, and Multi-Attack settings.}
            \label{fig:detectrl_indomain_f1}
        \end{subfigure}
    \label{fig:detectrl_indomain}
    \end{minipage}
    \hfill
    \begin{minipage}[t]{0.3\textwidth}
        \vspace{0pt}
        \centering

        \begin{subfigure}[t]{\linewidth}
            \centering
            \includegraphics[width=\linewidth]{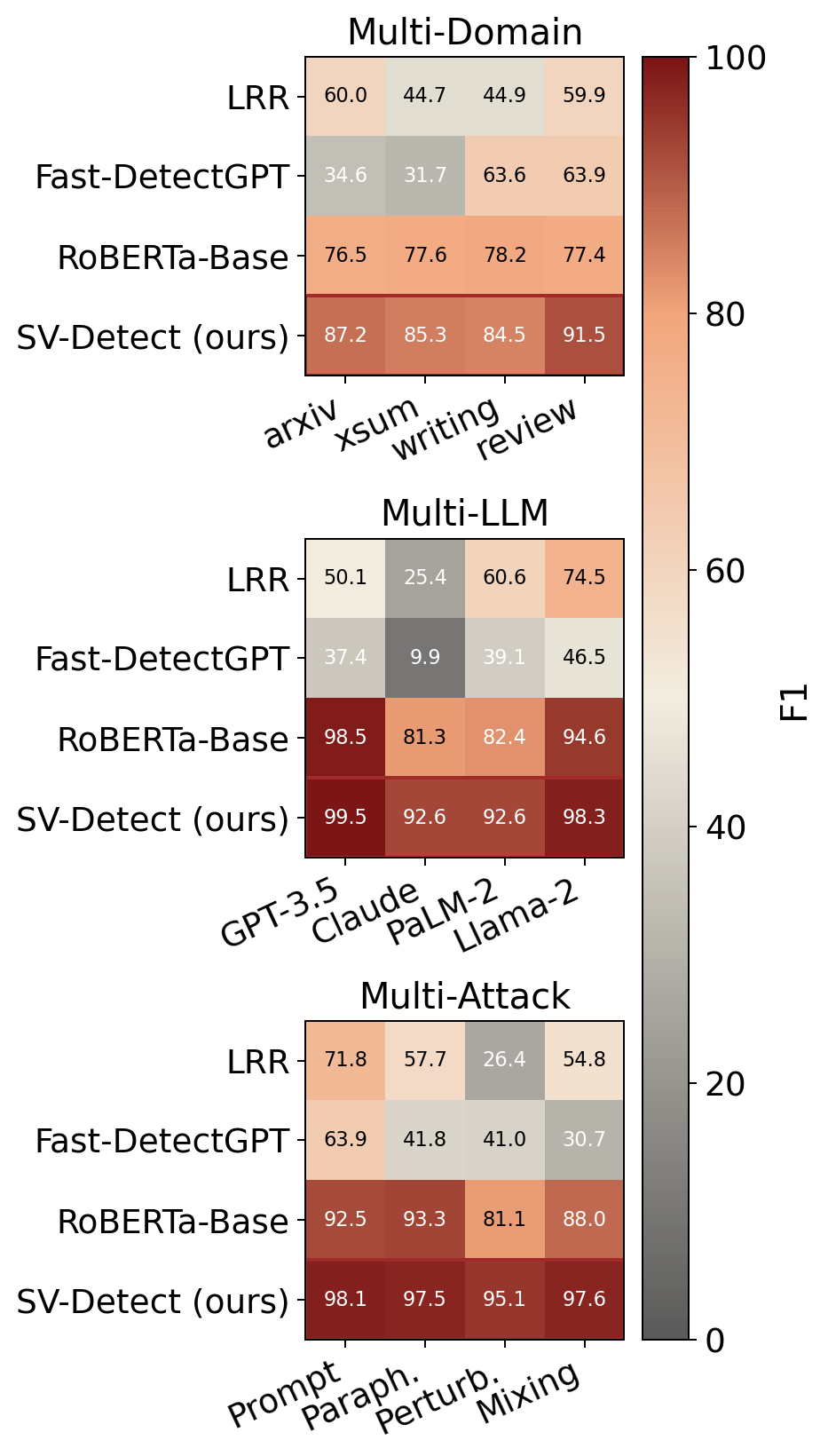}
            \caption{Cross-source generalization on DetectRL, per test subset $F_1$.}
            \label{fig:detectrl_crosssource}
        \end{subfigure}
    \end{minipage}

    \caption{Performance on DetectRL.}
    \label{fig:radars_combined}
    \vspace{-0.1cm}
\end{figure*}


\subsection{Experimental setup}
\label{sec:experiments_setup}

\noindent \textbf{Datasets.}
Our main evaluation uses three complementary benchmarks. Our primary benchmarks are \textbf{DetectRL}~\citep{DBLP:journals/corr/abs-2410-23746} and
\textbf{MIRAGE}~\citep{DBLP:journals/corr/abs-2509-14268}. DetectRL stresses robustness under distribution shifts across domains, source models, and attack families, while MIRAGE targets machine-writing scenarios, including generation, polishing, and rewriting.
We then assess cross-benchmark generalization by transferring detectors between \textbf{DetectRL} and \textbf{MIRAGE}, and by evaluating detectors trained on these benchmarks zero-shot on \textbf{PADBen}~\citep{padben}.
We additionally report results on the \textbf{COLING} dataset~\citep{DBLP:conf/coling/JawaharASL20} in Appendix Sec.~\ref{sec:coling}, and cross-benchmark transfer from DetectRL and MIRAGE to the \textbf{RAID} dataset~\citep{raid} in Appendix Sec.~\ref{sec:appendix_cross_benchmark}. We also use RAID for an additional matched-supervision comparison with RepreGuard~\citep{RepreGuard}, the closest representation-based detector in Sec.~\ref{sec:appendix_repreguard_vs_svdetect}.
\smallskip

\noindent \textbf{DetectRL evaluation details.}
We evaluate on the three settings provided by DetectRL:
(i) \textit{Multi-Domain}, spanning \texttt{ArXiv}, \texttt{XSum}, \texttt{Writing}, and \texttt{Review};
(ii) \textit{Multi-LLM}, spanning \texttt{GPT-3.5}, \texttt{Claude}, \texttt{PaLM-2}, and \texttt{Llama-2};
and (iii) \textit{Multi-Attack}, spanning prompt-based, paraphrase, perturbation, and data-mixing attacks.

For Multi-Domain and Multi-LLM, AI-generated examples are balanced across latent factors by subsampling each \((\texttt{llm\_type}, \texttt{data\_type})\) combination, using 50 AI-generated samples per combination for Multi-Domain and 200 for Multi-LLM. For Multi-Attack, both human-written and AI-generated training sets are subsampled to \(10{,}192\) examples. We always evaluate on the full benchmark test split.

Within each setting, we report both \textbf{same-source} and \textbf{cross-source} evaluation: same-source trains and tests on the same domain, source model, or attack family, while cross-source trains on one source and evaluates on another.
\smallskip

\noindent \textbf{MIRAGE evaluation details.}
We evaluate on MIRAGE under two settings, \textit{DIG} (Disjoint-Input Generation) and \textit{SIG} (Shared-Input Generation), and three transformations: \textsc{generate}, \textsc{polish}, and \textsc{rewrite}. The test splits are large-scale, e.g.\ \(16{,}411\) DIG-generate, \(14{,}776\) DIG-polish, \(15{,}735\) DIG-rewrite, \(16{,}388\) SIG-generate, \(14{,}776\) SIG-polish, and \(15{,}751\) SIG-rewrite examples.

We use the original training setup of MIRAGE that contains 500 human-written/machine-generated pairs, where the machine-generated side is GPT-3.5-Turbo-polished text. 
\smallskip

\noindent \textbf{PADBen evaluation details.}
We use additional benchmark, PADBen~\citep{padben}, purely as zero-shot cross-benchmark target: our MIRAGE- and DetectRL-trained detectors are applied directly to PADBen test splits without any retraining or threshold recalibration.

\emph{PADBen} targets paraphrase-based laundering and
human-edited AI-generated text. We report sentence-pair AUROC on the five benchmark regimes (\emph{paraphrase-source attribution},
\emph{general authorship detection}, \emph{AI-text laundering},
\emph{iterative paraphrase depth}, \emph{deep-paraphrase attack}).

\smallskip

\noindent \textbf{Evaluation protocol.} Following recent work~\citep{DBLP:journals/corr/abs-2509-14268,DBLP:journals/corr/abs-2412-10432,DBLP:journals/corr/abs-2310-05130,DBLP:journals/corr/abs-2301-11305}, for fair comparison with baseline methods, in all our main experiments, the reference model used for activation extraction is a frozen \texttt{GPT-Neo-2.7B}~\citep{Black2021GPTNeoLS}. For the RAID matched-supervision comparison against RepreGuard we additionally report results with a frozen \texttt{LLaMA-3.1-8B} backbone to match RepreGuard's original setup. We ablate choice of backbone model in Appendix~\ref{sec:ablation_backbone}.
Texts are tokenized with truncation to a maximum length of $2048$ tokens. For each text, we extract the mean-pooled hidden representation from every layer and compute cosine-similarity features with the corresponding steering vectors, as described in Section~\ref{sec:methodology}. We use mean pooling because it is simple, stable, and computationally cheap; richer summary statistics such as max pooling, attention-weighted pooling, or sentence-level aggregation are promising extensions but are left to future work. The downstream detector is trained on the same train data split used to construct the steering vectors.

Unless stated otherwise, the final detector is a pipeline consisting of \texttt{StandardScaler} followed by \texttt{LogisticRegression} with \texttt{liblinear} solver, \(\ell_2\)-regularization parameter $C=1.0$, and random seed $42$. For evaluation, we report AUROC, AUPR, TPR@FPR=5\%, Balanced Accuracy, Matthews correlation coefficient (MCC), and $F_1$.  
Full numerical tables corresponding to all result figures in this section are provided in Appendix Sec.~\ref{sec:thorough_tables}.

\subsection{Experimental results}

\noindent \textbf{DetectRL in-distribution performance.}
We follow \citep{DBLP:journals/corr/abs-2410-23746} and compare SV-Detect against both zero-shot and supervised baselines. In the same-source setting, the zero-shot baselines are Log-Likelihood~\citep{DBLP:journals/corr/abs-1908-09203}, Entropy~\citep{DBLP:conf/ecai/LavergneUY08}, Rank, Log-Rank~\citep{DBLP:journals/corr/abs-1906-04043}, LRR, NPR~\citep{DBLP:journals/corr/abs-2306-05540}, DetectGPT~\citep{DBLP:journals/corr/abs-2301-11305}, DNA-GPT~\citep{DBLP:journals/corr/abs-2305-17359}, Revise-Detect~\citep{DBLP:conf/emnlp/ZhuYCCFHDL0023}, Binoculars~\citep{DBLP:journals/corr/abs-2401-12070}, and Fast-DetectGPT~\citep{DBLP:journals/corr/abs-2310-05130}, while the supervised baselines are RoBERTa-Base~\citep{DBLP:journals/corr/abs-2105-09680}, RoBERTa-Large, XLM-RoBERTa-Base, and XLM-RoBERTa-Large~\citep{DBLP:journals/corr/abs-1911-02116}. We additionally add RepreGuard~\citep{RepreGuard} to the comparison, which is the closest supervised baseline to our work. For the fair comparison, we implemented RepreGuard using matched GPT-Neo-2.7B backbone based on the official provided code. Comparison of SV-Detect with RepreGuard using the default backbone of RepreGuard (LLaMA-3.1-8B-Instruct) is provided in Appendix Sec.~\ref{sec:appendix_repreguard_vs_svdetect}.

For readability, Fig.~\ref{fig:detectrl_indomain_auroc} and Fig.~\ref{fig:detectrl_indomain_f1} show a representative subset of baselines alongside SV-Detect: Log-Likelihood as a simple zero-shot scoring baseline, Binoculars as the strongest zero-shot detector, RoBERTa-Base as a strong supervised encoder baseline, and RepreGuard as the closest representation-based method. The full per-subset breakdown over all baselines is reported in Appendix Sec.~\ref{sec:thorough_tables}.

SV-Detect achieves near-perfect performance across all three DetectRL settings, with AUROC ranging between $99.8$ and $100.0$ and F1 between $98.8$ and $100.0$.


These results show that SV-Detect matches or slightly exceeds the strongest supervised baselines in the same-source setting, where performance is already close to saturated. At the same time, it substantially outperforms zero-shot and representation-based baselines, including Binoculars and RepreGuard, whose performance varies more strongly across domains, source models, and attack families.

\begin{figure*}[t]
      \centering
          \centering
          \includegraphics[width=\linewidth]{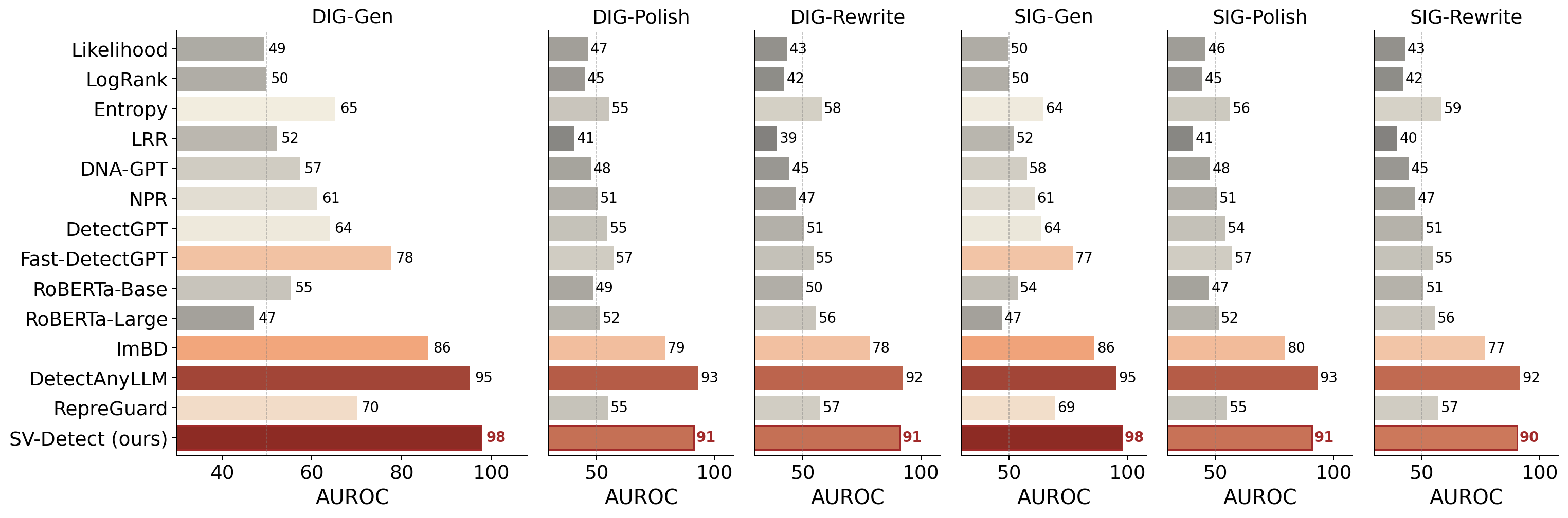}
          \caption{In-distribution performance on MIRAGE across the \textsc{Generate}, \textsc{Polish}, and \textsc{Rewrite} tasks under the DIG (left) and SIG (right) settings.}
          \label{fig:mirage_bars}
      \vspace{-0.1cm}
  \end{figure*}

\begin{figure}[t]
    \centering
    \includegraphics[width=\columnwidth]{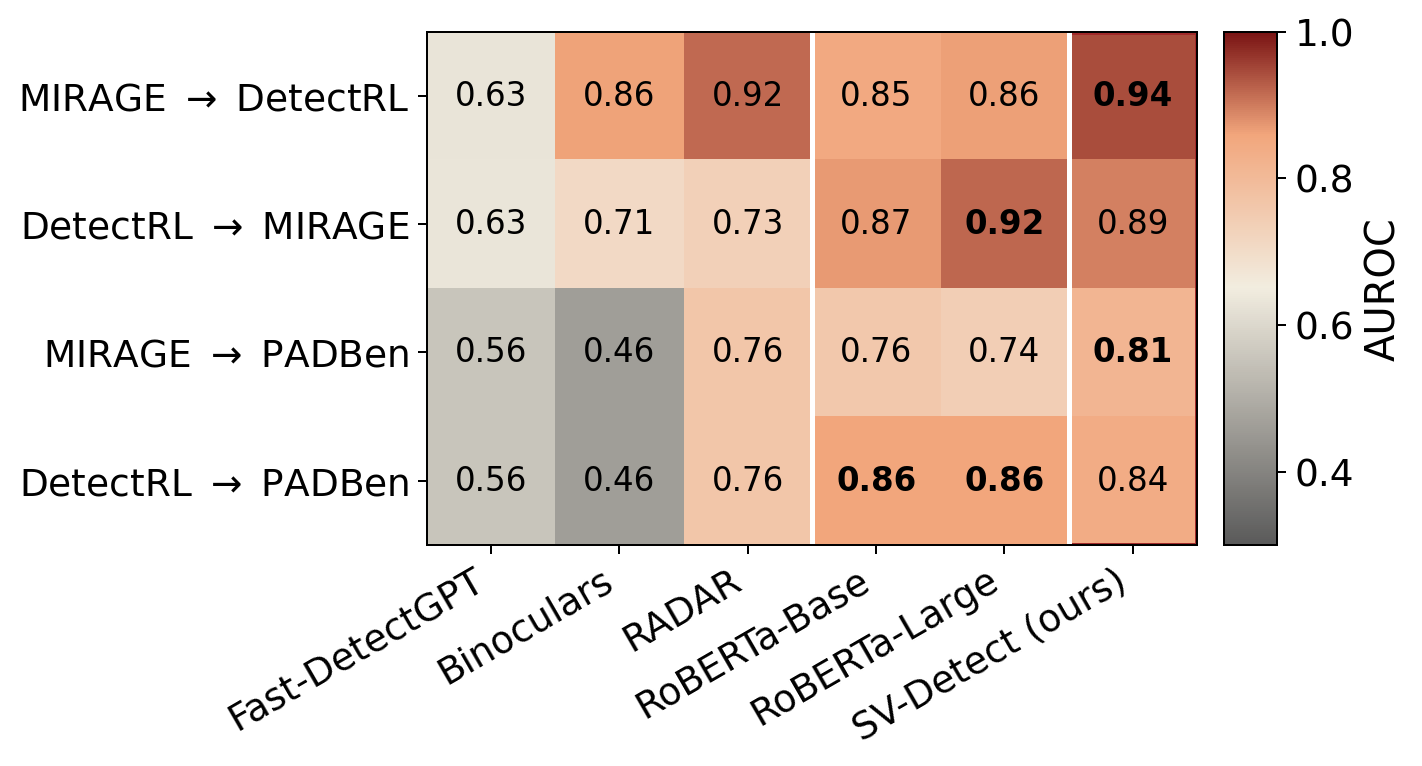}
    \caption{
    Cross-benchmark transfer. Grand-mean AUROC for each transfer direction
    (rows) and detector (columns)}
    \label{fig:cross_benchmark_matrix}
    \vspace{-0.1cm}
\end{figure}

\vspace{0.05cm}
\noindent \textbf{DetectRL cross-setting generalization.} We next evaluate cross-source generalization following \citep{DBLP:journals/corr/abs-2410-23746}. In this setting, a detector is trained on one source condition and evaluated on a different source condition within the same DetectRL setting. We compare against the baselines reported in \citep{DBLP:journals/corr/abs-2410-23746}: LRR and Fast-DetectGPT as zero-shot detectors, and RoBERTa-Base as a supervised baseline.

Fig.~\ref{fig:detectrl_crosssource} reports mean $F_1$ for each detector and evaluation subset, averaging over all train--test pairs in which the training subset differs from the evaluation subset. This isolates robustness to source mismatch: each column asks how well a detector performs on a target subset when it has not been trained on that subset.

SV-Detect achieves the strongest and most consistent cross-source performance across all three DetectRL settings. On \emph{Multi-Domain}, it improves over RoBERTa-Base by roughly $7$--$14$ $F_1$ points across target domains. On \emph{Multi-LLM}, it remains above $92$ $F_1$ for every target model and reaches near-perfect transfer on GPT-3.5 and Llama-2. On \emph{Multi-Attack}, it is similarly robust, maintaining at least $95$ $F_1$ across all attack families. In contrast, the zero-shot baselines vary substantially across target subsets, and RoBERTa-Base degrades more noticeably under source mismatch.

Overall, these cross-source results support the main claim of our approach: AI-generated text directions extracted from hidden representations encode a transferable human-vs-AI signal, enabling strong generalization across domains, source models, and attack families. Full train--test matrices are reported in Appendix Sec.~\ref{sec:thorough_tables}.


\smallskip
\begin{figure*}[t]
      \centering
      \begin{subfigure}[t]{0.49\textwidth}
          \centering
          \includegraphics[width=\linewidth]{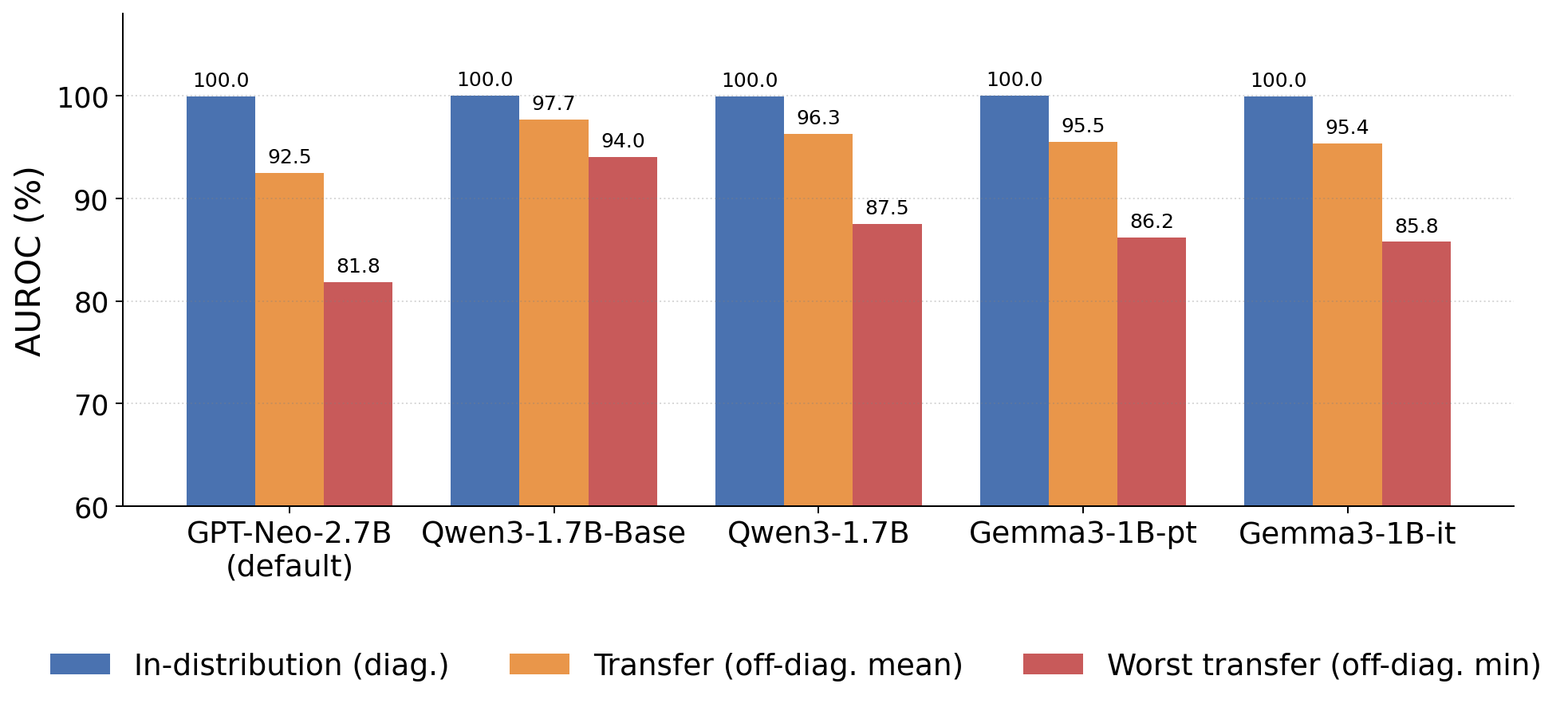}
          \caption{Backbone ablation}
          \label{fig:ablations_backbone}
      \end{subfigure}\hfill
      \begin{subfigure}[t]{0.49\textwidth}
          \centering
          \includegraphics[width=\linewidth]{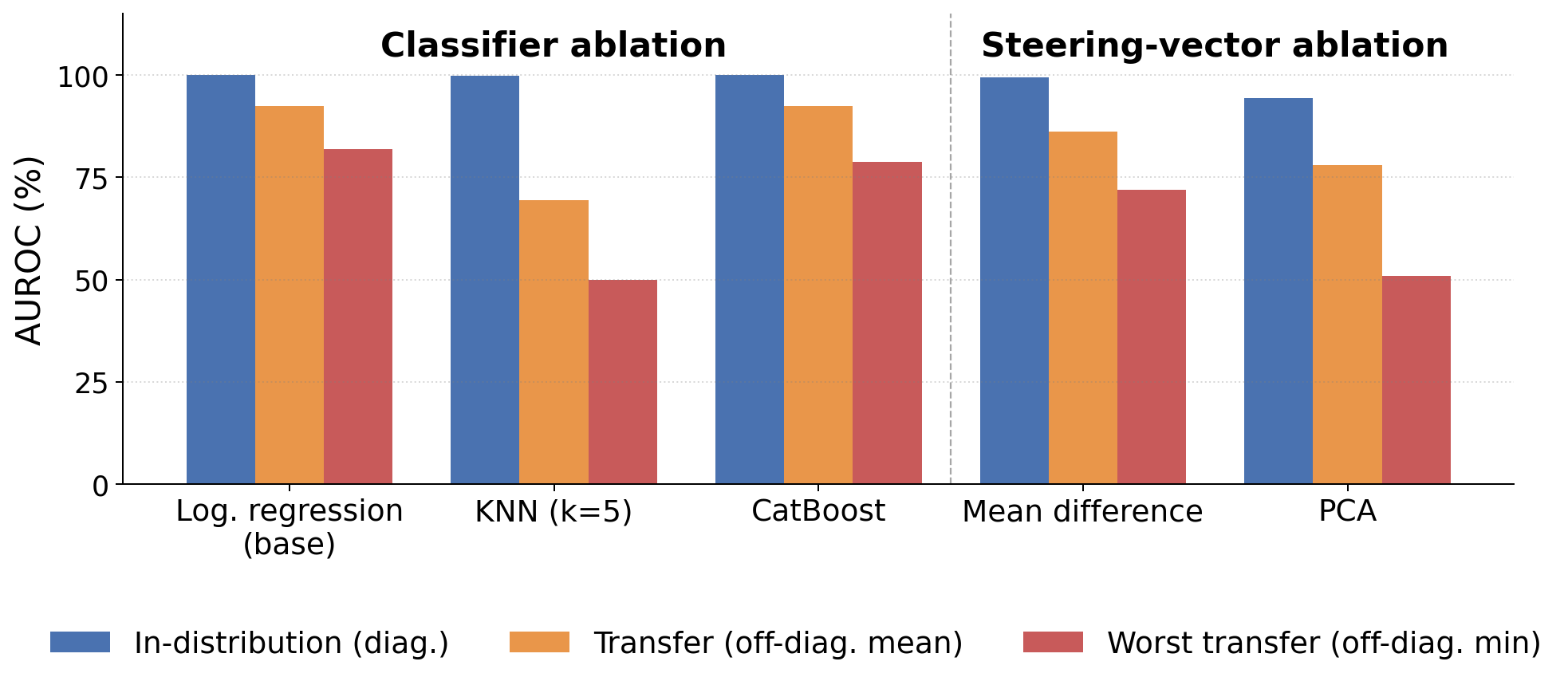}
          \caption{Classifier and construction of steering vector ablations}
          \label{fig:ablations}
      \end{subfigure}\hfill
      \caption{Ablation summary on DetectRL Multi-Domain transfer.}
      \label{fig:ablation_summary}
  \end{figure*}

\noindent \textbf{MIRAGE results.}
On MIRAGE, following \citep{DBLP:journals/corr/abs-2509-14268}, we compare SV-Detect against Log-Likelihood, LogRank, Entropy, RoBERTa-Base, RoBERTa-Large, LRR, DNA-GPT, NPR, DetectGPT, Fast-DetectGPT, ImBD~\citep{DBLP:journals/corr/abs-2412-10432}, and DetectAnyLLM~\citep{DBLP:journals/corr/abs-2509-14268}. As in the DetectRL experiments, we additionally include RepreGuard~\citep{RepreGuard} with a matched GPT-Neo-2.7B backbone as the closest representation-based baseline.

As seen on Fig.~\ref{fig:mirage_bars}, SV-Detect performs strongly across all six MIRAGE settings and achieves the best results on direct generation, with AUROC \(0.9777\) on DIG-\textsc{Generate} and \(0.9779\) on SIG-\textsc{Generate}. On \textsc{polish} and \textsc{rewrite}, it remains competitive but is outperformed by DetectAnyLLM.

In Appendix Sec.~\ref{sec:appendix_cross_benchmark}, we report an additional experiment on MIRAGE, where we use additional training data to construct separate task-specific per-layer steering directions corresponding to \textsc{generate}, \textsc{polish}, and \textsc{rewrite} subtasks. This variant substantially improves performance on the editing subtasks, showing that task-specific directions provide a richer representation of the different MIRAGE transformation types than the default \textsc{polish}-only direction.
\smallskip

\noindent \textbf{Cross-benchmark transfer.} We further analyze whether the representations learned by SV-Detect transfer across benchmark boundaries. Specifically, we train SV-Detect on one benchmark and evaluate it on a different target benchmark, without retraining, threshold calibration, or access to target-benchmark data.

We evaluate four transfer directions, using MIRAGE and DetectRL as source benchmarks and DetectRL, MIRAGE, and PADBen as targets: (i)~MIRAGE~$\to$~DetectRL, (ii)~DetectRL~$\to$~MIRAGE,
(iii)~MIRAGE~$\to$~PADBen, and (iv)~DetectRL~$\to$~PADBen.
In each case, the detector is trained only on the source benchmark's training pool described in Sec.~\ref{sec:experiments_setup}: for MIRAGE, we use the default 500 \textsc{Polish} training pairs, and for DetectRL, we fit a single steering direction per
layer using the union of the four DetectRL Multi-Domain training subsets. We evaluate on the target benchmark's test split without retraining, threshold calibration, or access to any target-benchmark data.

We compare against the strongest zero-shot detectors reported across the DetectRL and PADBen baselines: Fast-DetectGPT~\citep{DBLP:journals/corr/abs-2310-05130},
Binoculars~\citep{DBLP:journals/corr/abs-2401-12070}, and
RADAR~\citep{radar}. We also include RoBERTa-Base and RoBERTa-Large models fine-tuned on the same source pool as SV-Detect, giving a matched-supervision comparison with strong discriminative classifiers.
We report macro-averaged AUROC across the constituent settings of each target benchmark: 13 configurations for DetectRL
(4 Multi-Domain subsets, 4 Multi-LLM generators, and 5 Multi-Attack categories), 6 sentence-pair tasks for MIRAGE, and 5 sentence-pair regimes for PADBen.
Full per-cell breakdowns are provided in
Appendix Tables~\ref{tab:bidirectional_MD_full_appendix},
\ref{tab:detectrl_D_M_full_appendix}, and~\ref{tab:zeroshot_appendix}.

Figure~\ref{fig:cross_benchmark_matrix} summarizes the results. Overall, SV-Detect exhibits stable and consistently strong transfer across benchmarks: it is the best or second-best method in every transfer direction. The strongest results are obtained when MIRAGE is used as the source benchmark. SV-Detect is the top-performing method on both MIRAGE~$\to$~DetectRL and MIRAGE~$\to$~PADBen. When trained on DetectRL, SV-Detect also transfers well,
substantially outperforming all zero-shot detectors on both MIRAGE and PADBen. However, RoBERTa-Large trained on the same DetectRL pool performs slightly better in aggregate on these two targets.



\subsection{Ablation studies}
\label{sec:ablation}

\noindent \textbf{Ablation summary.}
Fig.~\ref{fig:ablations_backbone} and Fig.~\ref{fig:ablations} summarize three design choices in SV-Detect on DetectRL Multi-Domain transfer: choice of frozen LM backbone, the downstream classifier and the steering vector construction method. For each choice, we report three complementary quantities: mean in-distribution AUROC, mean transfer AUROC (averaged over off-diagonal pairs), and worst-case transfer AUROC (minimum over transfer pairs). This compact view captures not only average performance, but also robustness under the hardest distribution shifts.
\smallskip

\noindent \textbf{Choice of downstream classifier.}
Among the classifiers we consider, logistic regression is the most reliable overall and is therefore used as the default detector throughout the paper. While CatBoost remains competitive in-distribution, it is consistently weaker under transfer, especially in the worst case. KNN is substantially less stable and often collapses to near-chance performance on off-diagonal evaluations. This suggests that the main strength of SV-Detect lies in the steering-based representation itself rather than in a highly expressive nonlinear classification head.
\smallskip

\noindent \textbf{Choice of steering vector construction.}
The choice of steering vector construction has an even stronger effect. Logistic-regression-based steering vectors are clearly the most robust, achieving both the strongest average transfer and the strongest worst-case transfer. In contrast, mean-difference and PCA-based directions degrade much more severely, with worst-case transfer often approaching chance level. This shows that explicitly learning a discriminative direction in activation space is substantially more effective than relying on unsupervised variance directions or raw class-mean differences, and justifies our use of LogReg-based steering in the main experiments.
\smallskip

\noindent \textbf{Choice of frozen LM backbone.}
We use GPT-Neo as the default backbone in the main experiments for fair comparison with prior work. On Fig.~\ref{fig:ablations_backbone}, we compare several alternative backbones of similar scale. The results show that SV-Detect is not tied to GPT-Neo: all tested backbones perform strongly and, in fact, better than GPT-Neo at the worst-case transfer, with Qwen3-1.7B-Base being the most robust under transfer. Thus, backbone choice affects robustness, but the overall steering vector approach generalizes beyond the particular LM used in the main paper. More detailed analysis is provided in Appendix Sec.~\ref{sec:ablation_backbone}.

\section{What do the steering vectors detect?}
\label{sec:interpretation}

We study both \emph{where} SV-Detect extracts signal from and \emph{what} lexical or stylistic patterns are associated with the learned directions.

Our detector is a logistic regression on layer-wise projection scores,
\(\mathbf{s}(x) = (s_1(x), \dots, s_L(x))\),
with one feature per layer on DetectRL and one feature per layer--direction pair on MIRAGE. To identify the most important layers, we inspect the magnitude-weighted coefficients \(|w_l| \cdot \sigma_l\). On MIRAGE, the signal concentrates in the final layers (L29--31), and on DetectRL it is more distributed, with peaks at L0--2, L14--18, and L21. A more detailed layer-wise analysis is given in supplementary Section~\ref{sec:interpretation_suppl}.
\smallskip

\noindent \textbf{Logit-lens interpretation.}
\label{logit-lens}
For the top-contributing layers, we interpret the steering vector \(v_l\) by projecting it through the final layer norm and LM head:
\[
\mathrm{logits}_l = \texttt{LM\_head}(\texttt{LN}_f(v_l)).
\]
Top-ranked tokens correspond to the \(+v_l\) direction (AI-generated text side), and bottom-ranked tokens to \(-v_l\) (human-text side). This is a standard logit-lens probe applied to a learned direction.
\smallskip

\begin{figure}[t]
      \centering
      \begin{subfigure}{\columnwidth}
          \centering
          \includegraphics[width=\linewidth]{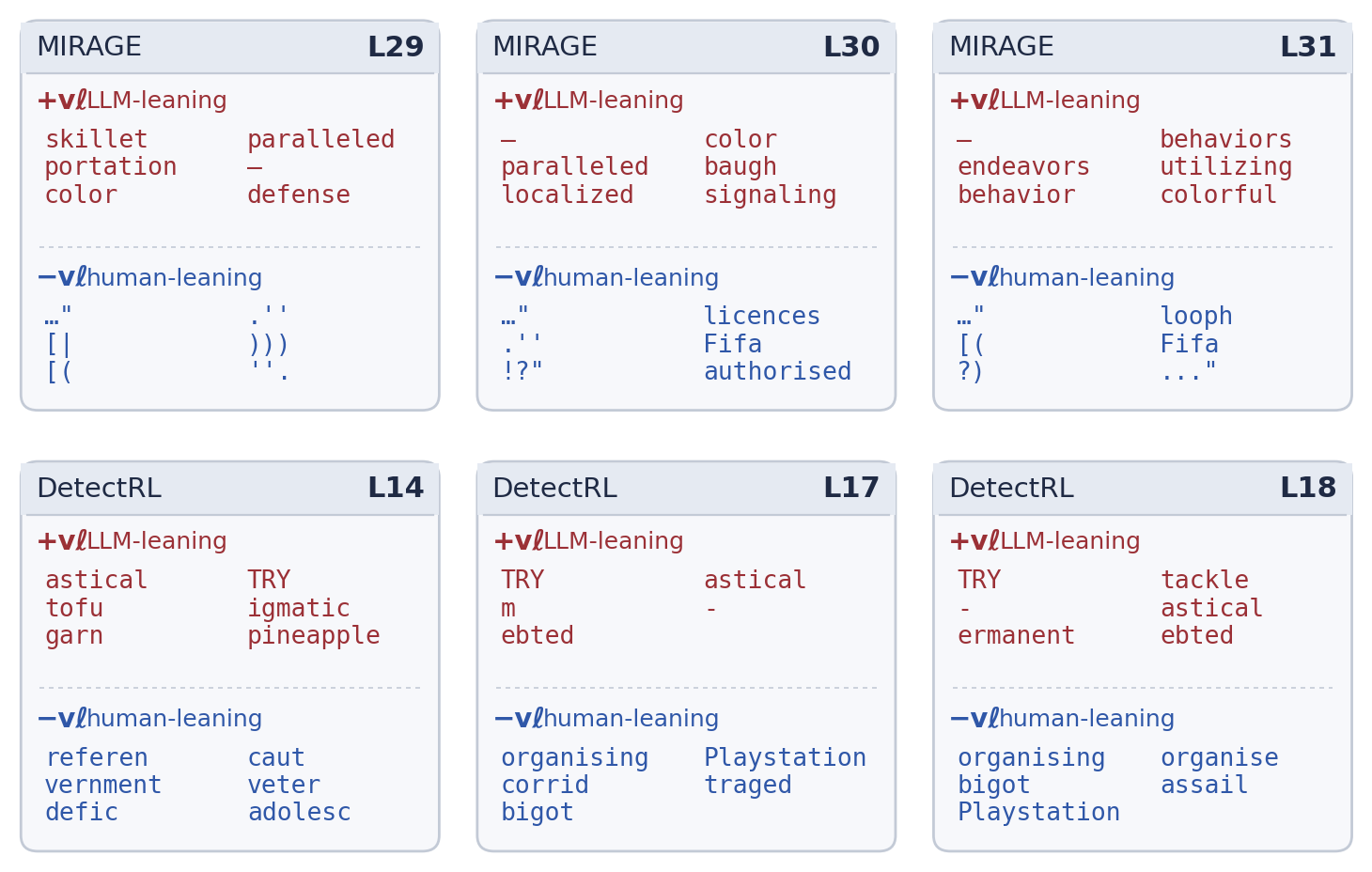}
          \caption{Top tokens at the most contributing layers.}
          \label{fig:logit_lens_tokens}
      \end{subfigure}\\[0.8em]
      \begin{subfigure}{\columnwidth}
          \centering
          \includegraphics[width=\linewidth]{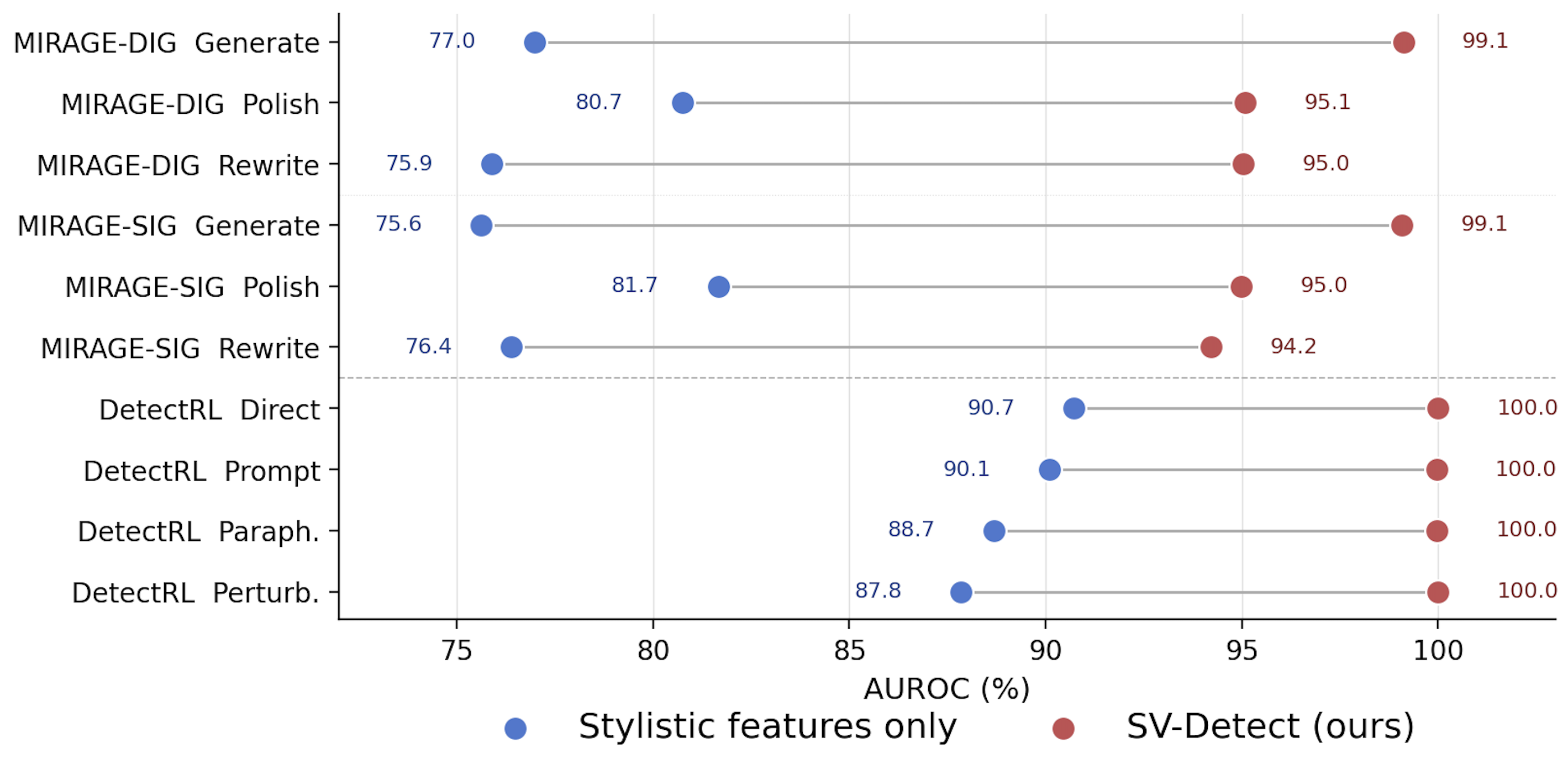}
          \caption{In-domain AUROC of a regex detector vs.\ the full pipeline.}
          \label{fig:interp_comparison}
      \end{subfigure}
      \caption{Steering vector interpretation.}
      \label{fig:interpretation_combined}
      \vspace{-0.3cm}
\end{figure}

\noindent \textbf{Lexical signatures.}
Figure~\ref{fig:logit_lens_tokens} shows that the learned directions are lexically interpretable, though differently on MIRAGE and DetectRL. On MIRAGE, the \(+v_l\) direction at L29-31 surfaces content-bearing, often technical or formal fragments, while the \(-v_l\) direction is dominated by punctuation-heavy fragments together with a few proper nouns and specific lexical items. On DetectRL, the \(+v_l\) direction at L14-18 reflects a more polished LLM-like register, while the \(-v_l\) direction surfaces more colloquial and topical fragments. Thus, the steering vectors align with meaningful lexical and stylistic differences between human-written and machine-generated text.
\smallskip

\noindent \textbf{Beyond surface cues.}
To quantify how much of the signal is explained by simple stylistic features, we train a logistic regression on interpretable regex-based counts derived from the logit-lens analysis.
The full setup is given in the Appendix Sec.~\ref{sec:interpretation_suppl}. These features achieve 76-82 AUROC on MIRAGE and 88-91 AUROC on DetectRL, but the full steering vector pipeline improves over them by 13-24 AUROC points across settings (Figure~\ref{fig:interp_comparison}). This suggests that the learned directions capture not only recognizable surface markers, but also additional signal in the hidden representations.
\smallskip

\begin{figure}[t]
    \centering
    \includegraphics[width=\columnwidth]{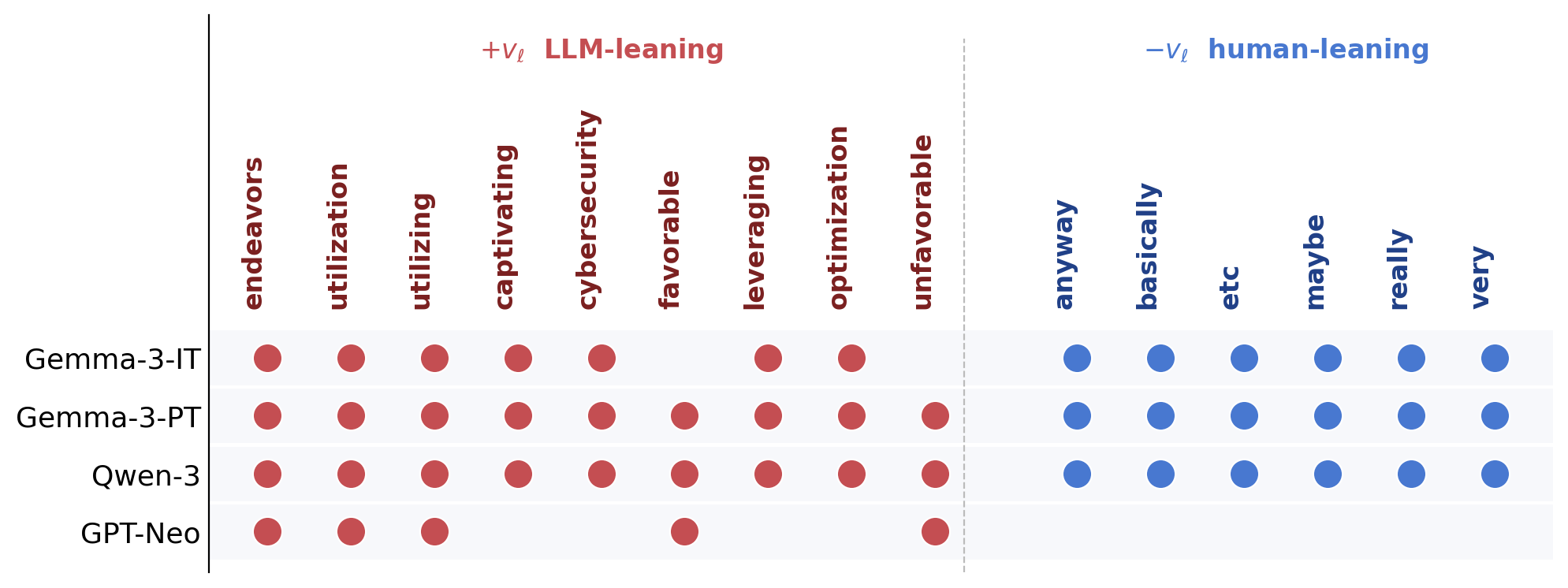}
    \caption{Consensus tokens across the four LMs. A colored dot indicates that the token appears in that LM's pooled top-token set.}
    \label{fig:cross_lm_consensus}
    \vspace{-0.3cm}
\end{figure}

\noindent \textbf{Cross-LM logit-lens analysis.}
We further test whether the lexical signal identified above is specific to GPT-Neo or shared across reference LMs. We retrain the steering vector pipeline on four LMs (GPT-Neo-2.7B, Qwen3-1.7B~\citep{yang2025qwen3technicalreport}, Gemma-3-1B-pt, and Gemma-3-1B-it~\citep{gemmateam2025gemma3technicalreport}) and decode the resulting dense steering vectors as in Sec.~\ref{sec:logit_lens}. Figure~\ref{fig:cross_lm_consensus} summarizes the result. Despite tokenizer- and corpus-specific differences, the same broad positive-side register recurs across models: all four surface \texttt{endeavors}, \texttt{utilization}, and \texttt{utilizing}, while the negative side is consistently more casual and discourse-like. This supports the view that SV-Detect captures a broader LLM-associated register rather than a GPT-Neo-specific artifact.

\smallskip
\noindent \textbf{Per-token contribution.}
SV-Detect also admits a token-level decomposition of the final score. For a fixed layer \(l\), we project each token activation at that layer onto the corresponding steering vector and use the resulting signed value as that token's contribution to the detector output. Figure~\ref{fig:teaser} visualizes this signal. LLM-generated text tends to exhibit localized spans with consistently strong positive evidence, whereas human-written text shows weaker and more heterogeneous contributions. This qualitative contrast is consistent with the analyses above.

\section{Conclusion}

We introduced \textbf{SV-Detect}, an AI-generated text detector based on steering vectors extracted from the hidden representations of a frozen language model. By representing each text through its alignment with layer-wise human-vs-AI directions and training lightweight classifier on top, SV-Detect provides a simple, interpretable alternative to text-level score-based and fully supervised detectors.

\smallskip
Experiments show that SV-Detect performs strongly both in-distribution and under challenging transfer settings, including shifts across domains, source models, attack families, and machine-editing transformations. It also transfers effectively across benchmarks, suggesting that it captures a benchmark-independent signal of AI-generated text.

\smallskip
Our interpretation analyses further show that the learned directions align with meaningful lexical and stylistic cues while also encoding substantial additional representation-level signal. Overall, these results support a representation-space view of AI-generated text detection: machine-generated text can be identified not only from its surface form, but also from stable directions in hidden-state space.
\section{Limitations and Future Work}

The proposed SV-Detect has several limitations. First, although it generalizes
well across domains, source models, attack families, and machine-editing
settings, it still depends on a probe LLM whose representation geometry affects
performance. Our backbone ablations in Section~\ref{sec:ablation_backbone}
show broadly consistent behavior across LM backbones, though some variation in
transfer robustness remains.
Second, while the method is much cheaper than perturbation-based detectors, it
still requires a full LLM forward pass and is therefore less efficient than
small encoder-based supervised baselines such as RoBERTa.
Third, our experiments focus on English benchmarks and on a fixed set of
contemporary generation and editing scenarios. Although MIRAGE and PADBen cover
important forms of machine-assisted writing, including polishing, rewriting,
paraphrasing, laundering, and human-edited AI-generated text, they do not fully
capture the broader spectrum of mixed human--AI authorship. In practice, many
documents may contain only localized AI-written passages, repeated human edits
over AI drafts, or iterative collaboration between a human writer and an LLM.
SV-Detect currently produces a document-level score and does not explicitly
attribute which spans are human-written, AI-generated, or jointly edited.
Finer-grained detection and attribution of mixed authorship therefore remain
important directions for future work.
Finally, SV-Detect remains a supervised detector. Although it generalizes better
than standard text classifiers in our experiments, it still relies on labeled
human/machine examples and may require periodic recalibration as generators,
prompting practices, editing tools, and human writing styles evolve. We do not
expect recalibration frequency to be fixed in advance: it should depend on the
deployment domain and be guided by held-out validation data from recent human
and AI-assisted writing. Since the probe LLM remains frozen, such adaptation
does not require fine-tuning the backbone; it only requires recomputing the
layer-wise directions and retraining the lightweight classifier on updated
examples.

These limitations point to several directions for future work. One natural
extension is to study SV-Detect in multilingual and cross-lingual settings,
where both the stylistic signal and the hidden-state geometry may differ
substantially. It would also be valuable to extend SV-Detect from document-level
classification to span-level or passage-level attribution for mixed human--AI
writing. Another important direction is to develop practical recalibration
protocols: for example, periodically refreshing the human/AI calibration set,
monitoring validation performance on recent writing samples, and updating the
directions when performance under distribution shift begins to degrade.
It would also be valuable to explore whether steering vector representations
can be combined with lighter-weight encoders to improve efficiency without
sacrificing robustness. More broadly, our results suggest that representation
space detection is a promising interface between robustness, interpretability,
and model internals, and future work could study whether similar ideas apply to
other modalities or to broader forms of synthetic content detection. Future work
could also compare LogReg-based/PCA/mean-difference steering to other supervised
linear projections such as shrinkage LDA or contrastive objectives.
\section*{Acknowledgments}
This work was supported by a Google DeepMind PhD Studentship, and the work utilized Queen Mary’s Andrena HPC facility, supported by QMUL Research-IT. This work was also supported by the Engineering and Physical Sciences Research Council [grant number EP/Y009800/1], through funding from Responsible Ai UK (KP0016). 
\bibliography{custom}

@article{repr_engineering,
  author       = {Andy Zou and
                  Long Phan and
                  Sarah Li Chen and
                  James Campbell and
                  Phillip Guo and
                  Richard Ren and
                  Alexander Pan and
                  Xuwang Yin and
                  Mantas Mazeika and
                  Ann{-}Kathrin Dombrowski and
                  Shashwat Goel and
                  Nathaniel Li and
                  Michael J. Byun and
                  Zifan Wang and
                  Alex Mallen and
                  Steven Basart and
                  Sanmi Koyejo and
                  Dawn Song and
                  Matt Fredrikson and
                  J. Zico Kolter and
                  Dan Hendrycks},
  title        = {Representation Engineering: {A} Top-Down Approach to {AI} Transparency},
  journal      = {CoRR},
  volume       = {abs/2310.01405},
  year         = {2023},
  url          = {https://doi.org/10.48550/arXiv.2310.01405},
  doi          = {10.48550/ARXIV.2310.01405},
  eprinttype   = {arXiv},
  eprint       = {2310.01405},
  bibsource    = {dblp computer science bibliography, https://dblp.org}
}

@article{DBLP:journals/csr/KehkashanRAAAHA25,
  author       = {Tanzila Kehkashan and
                  Raja Adil Riaz and
                  Ahmad Sami Al{-}Shamayleh and
                  Adnan Akhunzada and
                  Noman Ali and
                  Muhammad Hamza and
                  Faheem Akbar},
  title        = {AI-generated text detection: {A} comprehensive review of methods,
                  datasets, and applications},
  journal      = {Comput. Sci. Rev.},
  volume       = {58},
  pages        = {100793},
  year         = {2025},
  url          = {https://doi.org/10.1016/j.cosrev.2025.100793},
  doi          = {10.1016/J.COSREV.2025.100793},
  bibsource    = {dblp computer science bibliography, https://dblp.org}
}

@article{DBLP:journals/corr/abs-2410-23746,
  author       = {Junchao Wu and
                  Runzhe Zhan and
                  Derek F. Wong and
                  Shu Yang and
                  Xinyi Yang and
                  Yulin Yuan and
                  Lidia S. Chao},
  title        = {DetectRL: Benchmarking LLM-Generated Text Detection in Real-World
                  Scenarios},
  journal      = {CoRR},
  volume       = {abs/2410.23746},
  year         = {2024},
  url          = {https://doi.org/10.48550/arXiv.2410.23746},
  doi          = {10.48550/ARXIV.2410.23746},
  eprinttype   = {arXiv},
  eprint       = {2410.23746},
  bibsource    = {dblp computer science bibliography, https://dblp.org}
}

@article{DBLP:journals/corr/abs-2509-14268,
  author       = {Jiachen Fu and
                  Chun{-}Le Guo and
                  Chongyi Li},
  title        = {DetectAnyLLM: Towards Generalizable and Robust Detection of Machine-Generated
                  Text Across Domains and Models},
  journal      = {CoRR},
  volume       = {abs/2509.14268},
  year         = {2025},
  url          = {https://doi.org/10.48550/arXiv.2509.14268},
  doi          = {10.48550/ARXIV.2509.14268},
  eprinttype   = {arXiv},
  eprint       = {2509.14268},
  bibsource    = {dblp computer science bibliography, https://dblp.org}
}

@article{DBLP:journals/corr/abs-2501-11012,
  author       = {Yuxia Wang and
                  Artem Shelmanov and
                  Jonibek Mansurov and
                  Akim Tsvigun and
                  Vladislav Mikhailov and
                  Rui Xing and
                  Zhuohan Xie and
                  Jiahui Geng and
                  Giovanni Puccetti and
                  Ekaterina Artemova and
                  Jinyan Su and
                  Minh Ngoc Ta and
                  Mervat Abassy and
                  Kareem Ashraf Elozeiri and
                  Saad El Dine Ahmed El Etter and
                  Maiya Goloburda and
                  Tarek Mahmoud and
                  Raj Vardhan Tomar and
                  Nurkhan Laiyk and
                  Osama Mohammed Afzal and
                  Ryuto Koike and
                  Masahiro Kaneko and
                  Alham Fikri Aji and
                  Nizar Habash and
                  Iryna Gurevych and
                  Preslav Nakov},
  title        = {GenAI Content Detection Task 1: English and Multilingual Machine-Generated
                  Text Detection: {AI} vs. Human},
  journal      = {CoRR},
  volume       = {abs/2501.11012},
  year         = {2025},
  url          = {https://doi.org/10.48550/arXiv.2501.11012},
  doi          = {10.48550/ARXIV.2501.11012},
  eprinttype   = {arXiv},
  eprint       = {2501.11012},
  bibsource    = {dblp computer science bibliography, https://dblp.org}
}

@article{DBLP:journals/corr/abs-2412-10432,
  author       = {Jiaqi Chen and
                  Xiaoye Zhu and
                  Tianyang Liu and
                  Ying Chen and
                  Xinhui Chen and
                  Yiwen Yuan and
                  Chak Tou Leong and
                  Zuchao Li and
                  Tang Long and
                  Lei Zhang and
                  Chenyu Yan and
                  Guanghao Mei and
                  Jie Zhang and
                  Lefei Zhang},
  title        = {Imitate Before Detect: Aligning Machine Stylistic Preference for Machine-Revised
                  Text Detection},
  journal      = {CoRR},
  volume       = {abs/2412.10432},
  year         = {2024},
  url          = {https://doi.org/10.48550/arXiv.2412.10432},
  doi          = {10.48550/ARXIV.2412.10432},
  eprinttype   = {arXiv},
  eprint       = {2412.10432},
  bibsource    = {dblp computer science bibliography, https://dblp.org}
}

@article{DBLP:journals/corr/abs-2310-05130,
  author       = {Guangsheng Bao and
                  Yanbin Zhao and
                  Zhiyang Teng and
                  Linyi Yang and
                  Yue Zhang},
  title        = {Fast-DetectGPT: Efficient Zero-Shot Detection of Machine-Generated
                  Text via Conditional Probability Curvature},
  journal      = {CoRR},
  volume       = {abs/2310.05130},
  year         = {2023},
  url          = {https://doi.org/10.48550/arXiv.2310.05130},
  doi          = {10.48550/ARXIV.2310.05130},
  eprinttype   = {arXiv},
  eprint       = {2310.05130},
  bibsource    = {dblp computer science bibliography, https://dblp.org}
}

@article{DBLP:journals/corr/abs-2301-11305,
  author       = {Eric Mitchell and
                  Yoonho Lee and
                  Alexander Khazatsky and
                  Christopher D. Manning and
                  Chelsea Finn},
  title        = {DetectGPT: Zero-Shot Machine-Generated Text Detection using Probability
                  Curvature},
  journal      = {CoRR},
  volume       = {abs/2301.11305},
  year         = {2023},
  url          = {https://doi.org/10.48550/arXiv.2301.11305},
  doi          = {10.48550/ARXIV.2301.11305},
  eprinttype   = {arXiv},
  eprint       = {2301.11305},
  bibsource    = {dblp computer science bibliography, https://dblp.org}
}

@article{DBLP:journals/corr/abs-2401-12070,
  author       = {Abhimanyu Hans and
                  Avi Schwarzschild and
                  Valeriia Cherepanova and
                  Hamid Kazemi and
                  Aniruddha Saha and
                  Micah Goldblum and
                  Jonas Geiping and
                  Tom Goldstein},
  title        = {Spotting LLMs With Binoculars: Zero-Shot Detection of Machine-Generated
                  Text},
  journal      = {CoRR},
  volume       = {abs/2401.12070},
  year         = {2024},
  url          = {https://doi.org/10.48550/arXiv.2401.12070},
  doi          = {10.48550/ARXIV.2401.12070},
  eprinttype   = {arXiv},
  eprint       = {2401.12070},
  bibsource    = {dblp computer science bibliography, https://dblp.org}
}

@article{DBLP:journals/corr/abs-2305-17359,
  author       = {Xianjun Yang and
                  Wei Cheng and
                  Linda R. Petzold and
                  William Yang Wang and
                  Haifeng Chen},
  title        = {{DNA-GPT:} Divergent N-Gram Analysis for Training-Free Detection of
                  GPT-Generated Text},
  journal      = {CoRR},
  volume       = {abs/2305.17359},
  year         = {2023},
  url          = {https://doi.org/10.48550/arXiv.2305.17359},
  doi          = {10.48550/ARXIV.2305.17359},
  eprinttype   = {arXiv},
  eprint       = {2305.17359},
  bibsource    = {dblp computer science bibliography, https://dblp.org}
}

@article{DBLP:journals/corr/abs-2306-05540,
  author       = {Jinyan Su and
                  Terry Yue Zhuo and
                  Di Wang and
                  Preslav Nakov},
  title        = {DetectLLM: Leveraging Log Rank Information for Zero-Shot Detection
                  of Machine-Generated Text},
  journal      = {CoRR},
  volume       = {abs/2306.05540},
  year         = {2023},
  url          = {https://doi.org/10.48550/arXiv.2306.05540},
  doi          = {10.48550/ARXIV.2306.05540},
  eprinttype   = {arXiv},
  eprint       = {2306.05540},
  bibsource    = {dblp computer science bibliography, https://dblp.org}
}

@article{DBLP:journals/corr/abs-2109-04825,
  author       = {Laida Kushnareva and
                  Daniil Cherniavskii and
                  Vladislav Mikhailov and
                  Ekaterina Artemova and
                  Serguei Barannikov and
                  Alexander Bernstein and
                  Irina Piontkovskaya and
                  Dmitri Piontkovski and
                  Evgeny Burnaev},
  title        = {Artificial Text Detection via Examining the Topology of Attention
                  Maps},
  journal      = {CoRR},
  volume       = {abs/2109.04825},
  year         = {2021},
  url          = {https://arxiv.org/abs/2109.04825},
  eprinttype   = {arXiv},
  eprint       = {2109.04825},
  bibsource    = {dblp computer science bibliography, https://dblp.org}
  }

@article{DBLP:journals/corr/abs-2410-08113,
  author       = {Kristian Kuznetsov and
                  Eduard Tulchinskii and
                  Laida Kushnareva and
                  German Magai and
                  Serguei Barannikov and
                  Sergey I. Nikolenko and
                  Irina Piontkovskaya},
  title        = {Robust AI-Generated Text Detection by Restricted Embeddings},
  journal      = {CoRR},
  volume       = {abs/2410.08113},
  year         = {2024},
  url          = {https://doi.org/10.48550/arXiv.2410.08113},
  doi          = {10.48550/ARXIV.2410.08113},
  eprinttype   = {arXiv},
  eprint       = {2410.08113},
  bibsource    = {dblp computer science bibliography, https://dblp.org}
}

@article{WuYZYCW25,
  author       = {Junchao Wu and
                  Shu Yang and
                  Runzhe Zhan and
                  Yulin Yuan and
                  Lidia S. Chao and
                  Derek Fai Wong},
  title        = {A Survey on LLM-Generated Text Detection: Necessity, Methods, and
                  Future Directions},
  journal      = {Comput. Linguistics},
  volume       = {51},
  number       = {1},
  pages        = {275--338},
  year         = {2025},
  url          = {https://doi.org/10.1162/coli\_a\_00549},
  doi          = {10.1162/COLI\_A\_00549},
  bibsource    = {dblp computer science bibliography, https://dblp.org}
}

@article{DBLP:journals/corr/abs-2306-15666,
  author       = {Debora Weber{-}Wulff and
                  Alla Anohina{-}Naumeca and
                  Sonja Bjelobaba and
                  Tom{\'{a}}s Folt{\'{y}}nek and
                  Jean Guerrero{-}Dib and
                  Olumide Popoola and
                  Petr Sigut and
                  Lorna Waddington},
  title        = {Testing of Detection Tools for AI-Generated Text},
  journal      = {CoRR},
  volume       = {abs/2306.15666},
  year         = {2023},
  url          = {https://doi.org/10.48550/arXiv.2306.15666},
  doi          = {10.48550/ARXIV.2306.15666},
  eprinttype   = {arXiv},
  eprint       = {2306.15666},
  bibsource    = {dblp computer science bibliography, https://dblp.org}
}

@inproceedings{DBLP:conf/coling/JawaharASL20,
  author       = {Ganesh Jawahar and
                  Muhammad Abdul{-}Mageed and
                  Laks V. S. Lakshmanan},
  editor       = {Donia Scott and
                  N{\'{u}}ria Bel and
                  Chengqing Zong},
  title        = {Automatic Detection of Machine Generated Text: {A} Critical Survey},
  booktitle    = {Proceedings of the 28th International Conference on Computational
                  Linguistics, {COLING} 2020, Barcelona, Spain (Online), December 8-13,
                  2020},
  pages        = {2296--2309},
  publisher    = {International Committee on Computational Linguistics},
  year         = {2020},
  url          = {https://doi.org/10.18653/v1/2020.coling-main.208},
  doi          = {10.18653/V1/2020.COLING-MAIN.208},
  bibsource    = {dblp computer science bibliography, https://dblp.org}
}

@inproceedings{DBLP:conf/acl/LiLCBWWY0024,
  author       = {Yafu Li and
                  Qintong Li and
                  Leyang Cui and
                  Wei Bi and
                  Zhilin Wang and
                  Longyue Wang and
                  Linyi Yang and
                  Shuming Shi and
                  Yue Zhang},
  editor       = {Lun{-}Wei Ku and
                  Andre Martins and
                  Vivek Srikumar},
  title        = {{MAGE:} Machine-generated Text Detection in the Wild},
  booktitle    = {Proceedings of the 62nd Annual Meeting of the Association for Computational
                  Linguistics (Volume 1: Long Papers), {ACL} 2024, Bangkok, Thailand,
                  August 11-16, 2024},
  pages        = {36--53},
  publisher    = {Association for Computational Linguistics},
  year         = {2024},
  url          = {https://doi.org/10.18653/v1/2024.acl-long.3},
  doi          = {10.18653/V1/2024.ACL-LONG.3},
  bibsource    = {dblp computer science bibliography, https://dblp.org}
}

@inproceedings{DBLP:conf/naacl/TuftsZL25,
  author       = {Brian Tufts and
                  Xuandong Zhao and
                  Lei Li},
  editor       = {Luis Chiruzzo and
                  Alan Ritter and
                  Lu Wang},
  title        = {A Practical Examination of AI-Generated Text Detectors for Large Language
                  Models},
  booktitle    = {Findings of the Association for Computational Linguistics: {NAACL}
                  2025, Albuquerque, New Mexico, USA, April 29 - May 4, 2025},
  series       = {Findings of {ACL}},
  pages        = {4824--4841},
  publisher    = {Association for Computational Linguistics},
  year         = {2025},
  url          = {https://doi.org/10.18653/v1/2025.findings-naacl.271},
  doi          = {10.18653/V1/2025.FINDINGS-NAACL.271},
  bibsource    = {dblp computer science bibliography, https://dblp.org}
}

@misc{kushnareva2024aigeneratedtextboundarydetection,
      title={AI-generated text boundary detection with RoFT}, 
      author={Laida Kushnareva and Tatiana Gaintseva and German Magai and Serguei Barannikov and Dmitry Abulkhanov and Kristian Kuznetsov and Eduard Tulchinskii and Irina Piontkovskaya and Sergey Nikolenko},
      year={2024},
      eprint={2311.08349},
      archivePrefix={arXiv},
      primaryClass={cs.CL},
      url={https://arxiv.org/abs/2311.08349}, 
}

@misc{nostalgebraist2020,
  author = {nostalgebraist},
  title = {interpreting GPT: the logit lens},
  year = {2020},
  month = aug,
  howpublished = {\url{https://www.lesswrong.com/posts/AcKRB8wDpdaN6v6ru/interpreting-gpt-the-logit-lens}},
  note = {LessWrong post, accessed 2026-05-25}
}

@article{DBLP:journals/corr/abs-1908-09203,
  author       = {Irene Solaiman and
                  Miles Brundage and
                  Jack Clark and
                  Amanda Askell and
                  Ariel Herbert{-}Voss and
                  Jeff Wu and
                  Alec Radford and
                  Jasmine Wang},
  title        = {Release Strategies and the Social Impacts of Language Models},
  journal      = {CoRR},
  volume       = {abs/1908.09203},
  year         = {2019},
  url          = {http://arxiv.org/abs/1908.09203},
  eprinttype   = {arXiv},
  eprint       = {1908.09203},
  bibsource    = {dblp computer science bibliography, https://dblp.org}
}

@inproceedings{DBLP:conf/ecai/LavergneUY08,
  author       = {Thomas Lavergne and
                  Tanguy Urvoy and
                  Fran{\c{c}}ois Yvon},
  editor       = {Benno Stein and
                  Efstathios Stamatatos and
                  Moshe Koppel},
  title        = {Detecting Fake Content with Relative Entropy Scoring},
  booktitle    = {Proceedings of the ECAI'08 Workshop on Uncovering Plagiarism, Authorship
                  and Social Software Misuse, Patras, Greece, July 22, 2008},
  series       = {{CEUR} Workshop Proceedings},
  publisher    = {CEUR-WS.org},
  year         = {2008},
  url          = {https://ceur-ws.org/Vol-377/paper4.pdf},
  bibsource    = {dblp computer science bibliography, https://dblp.org}
}

@article{DBLP:journals/corr/abs-1906-04043,
  author       = {Sebastian Gehrmann and
                  Hendrik Strobelt and
                  Alexander M. Rush},
  title        = {{GLTR:} Statistical Detection and Visualization of Generated Text},
  journal      = {CoRR},
  volume       = {abs/1906.04043},
  year         = {2019},
  url          = {http://arxiv.org/abs/1906.04043},
  eprinttype   = {arXiv},
  eprint       = {1906.04043},
  bibsource    = {dblp computer science bibliography, https://dblp.org}
}

@inproceedings{DBLP:conf/emnlp/ZhuYCCFHDL0023,
  author       = {Biru Zhu and
                  Lifan Yuan and
                  Ganqu Cui and
                  Yangyi Chen and
                  Chong Fu and
                  Bingxiang He and
                  Yangdong Deng and
                  Zhiyuan Liu and
                  Maosong Sun and
                  Ming Gu},
  editor       = {Houda Bouamor and
                  Juan Pino and
                  Kalika Bali},
  title        = {Beat LLMs at Their Own Game: Zero-Shot LLM-Generated Text Detection
                  via Querying ChatGPT},
  booktitle    = {Proceedings of the 2023 Conference on Empirical Methods in Natural
                  Language Processing, {EMNLP} 2023, Singapore, December 6-10, 2023},
  pages        = {7470--7483},
  publisher    = {Association for Computational Linguistics},
  year         = {2023},
  url          = {https://doi.org/10.18653/v1/2023.emnlp-main.463},
  doi          = {10.18653/V1/2023.EMNLP-MAIN.463},
  bibsource    = {dblp computer science bibliography, https://dblp.org}
}

@article{DBLP:journals/corr/abs-2105-09680,
  author       = {Sungjoon Park and
                  Jihyung Moon and
                  Sungdong Kim and
                  Won{-}Ik Cho and
                  Jiyoon Han and
                  Jangwon Park and
                  Chisung Song and
                  Junseong Kim and
                  Yongsook Song and
                  Tae Hwan Oh and
                  Joohong Lee and
                  Juhyun Oh and
                  Sungwon Lyu and
                  Younghoon Jeong and
                  Inkwon Lee and
                  Sangwoo Seo and
                  Dongjun Lee and
                  Hyunwoo Kim and
                  Myeonghwa Lee and
                  Seongbo Jang and
                  Seungwon Do and
                  Sunkyoung Kim and
                  Kyungtae Lim and
                  Jongwon Lee and
                  Kyumin Park and
                  Jamin Shin and
                  Seonghyun Kim and
                  Eunjeong Lucy Park and
                  Alice Oh and
                  Jung{-}Woo Ha and
                  Kyunghyun Cho},
  title        = {{KLUE:} Korean Language Understanding Evaluation},
  journal      = {CoRR},
  volume       = {abs/2105.09680},
  year         = {2021},
  url          = {https://arxiv.org/abs/2105.09680},
  eprinttype   = {arXiv},
  eprint       = {2105.09680},
  bibsource    = {dblp computer science bibliography, https://dblp.org}
}

@article{DBLP:journals/corr/abs-1911-02116,
  author       = {Alexis Conneau and
                  Kartikay Khandelwal and
                  Naman Goyal and
                  Vishrav Chaudhary and
                  Guillaume Wenzek and
                  Francisco Guzm{\'{a}}n and
                  Edouard Grave and
                  Myle Ott and
                  Luke Zettlemoyer and
                  Veselin Stoyanov},
  title        = {Unsupervised Cross-lingual Representation Learning at Scale},
  journal      = {CoRR},
  volume       = {abs/1911.02116},
  year         = {2019},
  url          = {http://arxiv.org/abs/1911.02116},
  eprinttype   = {arXiv},
  eprint       = {1911.02116},
  bibsource    = {dblp computer science bibliography, https://dblp.org}
}

@misc{Black2021GPTNeoLS,
  title={GPT-Neo: Large Scale Autoregressive Language Modeling with Mesh-Tensorflow},
  author={Sid Black and Leo Gao and Phil Wang and Connor Leahy and Stella Biderman},
  year={2021},
  howpublished={\url{https://github.com/EleutherAI/gpt-neo}}
}

@misc{yang2025qwen3technicalreport,
      title={Qwen3 Technical Report}, 
      author={An Yang and Anfeng Li and Baosong Yang and Beichen Zhang and Binyuan Hui and Bo Zheng and Bowen Yu and Chang Gao and Chengen Huang and Chenxu Lv and Chujie Zheng and Dayiheng Liu and Fan Zhou and Fei Huang and Feng Hu and Hao Ge and Haoran Wei and Huan Lin and Jialong Tang and Jian Yang and Jianhong Tu and Jianwei Zhang and Jianxin Yang and Jiaxi Yang and Jing Zhou and Jingren Zhou and Junyang Lin and Kai Dang and Keqin Bao and Kexin Yang and Le Yu and Lianghao Deng and Mei Li and Mingfeng Xue and Mingze Li and Pei Zhang and Peng Wang and Qin Zhu and Rui Men and Ruize Gao and Shixuan Liu and Shuang Luo and Tianhao Li and Tianyi Tang and Wenbiao Yin and Xingzhang Ren and Xinyu Wang and Xinyu Zhang and Xuancheng Ren and Yang Fan and Yang Su and Yichang Zhang and Yinger Zhang and Yu Wan and Yuqiong Liu and Zekun Wang and Zeyu Cui and Zhenru Zhang and Zhipeng Zhou and Zihan Qiu},
      year={2025},
      eprint={2505.09388},
      archivePrefix={arXiv},
      primaryClass={cs.CL},
      url={https://arxiv.org/abs/2505.09388}, 
}

@misc{gemmateam2025gemma3technicalreport,
      title={Gemma 3 Technical Report}, 
      author={Gemma Team and Aishwarya Kamath and Johan Ferret and Shreya Pathak and Nino Vieillard and Ramona Merhej and Sarah Perrin and Tatiana Matejovicova and Alexandre Ramé and Morgane Rivière and Louis Rouillard and Thomas Mesnard and Geoffrey Cideron and Jean-bastien Grill and Sabela Ramos and Edouard Yvinec and Michelle Casbon and Etienne Pot and Ivo Penchev and Gaël Liu and Francesco Visin and Kathleen Kenealy and Lucas Beyer and Xiaohai Zhai and Anton Tsitsulin and Robert Busa-Fekete and Alex Feng and Noveen Sachdeva and Benjamin Coleman and Yi Gao and Basil Mustafa and Iain Barr and Emilio Parisotto and David Tian and Matan Eyal and Colin Cherry and Jan-Thorsten Peter and Danila Sinopalnikov and Surya Bhupatiraju and Rishabh Agarwal and Mehran Kazemi and Dan Malkin and Ravin Kumar and David Vilar and Idan Brusilovsky and Jiaming Luo and Andreas Steiner and Abe Friesen and Abhanshu Sharma and Abheesht Sharma and Adi Mayrav Gilady and Adrian Goedeckemeyer and Alaa Saade and Alex Feng and Alexander Kolesnikov and Alexei Bendebury and Alvin Abdagic and Amit Vadi and András György and André Susano Pinto and Anil Das and Ankur Bapna and Antoine Miech and Antoine Yang and Antonia Paterson and Ashish Shenoy and Ayan Chakrabarti and Bilal Piot and Bo Wu and Bobak Shahriari and Bryce Petrini and Charlie Chen and Charline Le Lan and Christopher A. Choquette-Choo and CJ Carey and Cormac Brick and Daniel Deutsch and Danielle Eisenbud and Dee Cattle and Derek Cheng and Dimitris Paparas and Divyashree Shivakumar Sreepathihalli and Doug Reid and Dustin Tran and Dustin Zelle and Eric Noland and Erwin Huizenga and Eugene Kharitonov and Frederick Liu and Gagik Amirkhanyan and Glenn Cameron and Hadi Hashemi and Hanna Klimczak-Plucińska and Harman Singh and Harsh Mehta and Harshal Tushar Lehri and Hussein Hazimeh and Ian Ballantyne and Idan Szpektor and Ivan Nardini and Jean Pouget-Abadie and Jetha Chan and Joe Stanton and John Wieting and Jonathan Lai and Jordi Orbay and Joseph Fernandez and Josh Newlan and Ju-yeong Ji and Jyotinder Singh and Kat Black and Kathy Yu and Kevin Hui and Kiran Vodrahalli and Klaus Greff and Linhai Qiu and Marcella Valentine and Marina Coelho and Marvin Ritter and Matt Hoffman and Matthew Watson and Mayank Chaturvedi and Michael Moynihan and Min Ma and Nabila Babar and Natasha Noy and Nathan Byrd and Nick Roy and Nikola Momchev and Nilay Chauhan and Noveen Sachdeva and Oskar Bunyan and Pankil Botarda and Paul Caron and Paul Kishan Rubenstein and Phil Culliton and Philipp Schmid and Pier Giuseppe Sessa and Pingmei Xu and Piotr Stanczyk and Pouya Tafti and Rakesh Shivanna and Renjie Wu and Renke Pan and Reza Rokni and Rob Willoughby and Rohith Vallu and Ryan Mullins and Sammy Jerome and Sara Smoot and Sertan Girgin and Shariq Iqbal and Shashir Reddy and Shruti Sheth and Siim Põder and Sijal Bhatnagar and Sindhu Raghuram Panyam and Sivan Eiger and Susan Zhang and Tianqi Liu and Trevor Yacovone and Tyler Liechty and Uday Kalra and Utku Evci and Vedant Misra and Vincent Roseberry and Vlad Feinberg and Vlad Kolesnikov and Woohyun Han and Woosuk Kwon and Xi Chen and Yinlam Chow and Yuvein Zhu and Zichuan Wei and Zoltan Egyed and Victor Cotruta and Minh Giang and Phoebe Kirk and Anand Rao and Kat Black and Nabila Babar and Jessica Lo and Erica Moreira and Luiz Gustavo Martins and Omar Sanseviero and Lucas Gonzalez and Zach Gleicher and Tris Warkentin and Vahab Mirrokni and Evan Senter and Eli Collins and Joelle Barral and Zoubin Ghahramani and Raia Hadsell and Yossi Matias and D. Sculley and Slav Petrov and Noah Fiedel and Noam Shazeer and Oriol Vinyals and Jeff Dean and Demis Hassabis and Koray Kavukcuoglu and Clement Farabet and Elena Buchatskaya and Jean-Baptiste Alayrac and Rohan Anil and Dmitry and Lepikhin and Sebastian Borgeaud and Olivier Bachem and Armand Joulin and Alek Andreev and Cassidy Hardin and Robert Dadashi and Léonard Hussenot},
      year={2025},
      eprint={2503.19786},
      archivePrefix={arXiv},
      primaryClass={cs.CL},
      url={https://arxiv.org/abs/2503.19786}, 
}

@inproceedings{raid,
  author       = {Liam Dugan and
                  Alyssa Hwang and
                  Filip Trhl{\'{\i}}k and
                  Andrew Zhu and
                  Josh Magnus Ludan and
                  Hainiu Xu and
                  Daphne Ippolito and
                  Chris Callison{-}Burch},
  editor       = {Lun{-}Wei Ku and
                  Andre Martins and
                  Vivek Srikumar},
  title        = {{RAID:} {A} Shared Benchmark for Robust Evaluation of Machine-Generated
                  Text Detectors},
  booktitle    = {Proceedings of the 62nd Annual Meeting of the Association for Computational
                  Linguistics (Volume 1: Long Papers), {ACL} 2024, Bangkok, Thailand,
                  August 11-16, 2024},
  year         = {2024},
}

@article{padben,
  author       = {Yiwei Zha and
                  Rui Min and
                  Shanu Sushmita},
  title        = {PADBen: {A} Comprehensive Benchmark for Evaluating {AI} Text Detectors
                  Against Paraphrase Attacks},
  journal      = {CoRR},
  volume       = {abs/2511.00416},
  year         = {2025},
}

@article{RepreGuard,
  author       = {Xin Chen and
                  Junchao Wu and
                  Shu Yang and
                  Runzhe Zhan and
                  Zeyu Wu and
                  Ziyang Luo and
                  Di Wang and
                  Min Yang and
                  Lidia S. Chao and
                  Derek F. Wong},
  title        = {RepreGuard: Detecting LLM-Generated Text by Revealing Hidden Representation
                  Patterns},
  journal      = {Trans. Assoc. Comput. Linguistics},
  volume       = {13},
  year         = {2025},
}

@article{radar,
  author       = {Xiaomeng Hu and
                  Pin{-}Yu Chen and
                  Tsung{-}Yi Ho},
  title        = {{RADAR:} Robust AI-Text Detection via Adversarial Learning},
  journal      = {CoRR},
  volume       = {abs/2307.03838},
  year         = {2023},
  url          = {https://doi.org/10.48550/arXiv.2307.03838},
  doi          = {10.48550/ARXIV.2307.03838},
  eprinttype   = {arXiv},
  eprint       = {2307.03838},
  bibsource    = {dblp computer science bibliography, https://dblp.org}
}

\clearpage
\appendix
\tableofcontents

\section{Potential Risks}
Like other AI-generated text detectors, SV-Detect could be misused in high-stakes settings such as academic misconduct accusations, moderation, or authorship disputes if its predictions are treated as definitive evidence rather than probabilistic signals. False positives may unfairly penalize human authors, while false negatives may allow machine-generated text to evade detection. We therefore view SV-Detect as a decision-support tool rather than a standalone adjudicator.

A second risk is adversarial adaptation: public knowledge of the detector may encourage authors or model providers to modify generation style or post-process outputs in ways that reduce detectability. Although our experiments show robustness across several shifts and editing settings, we do not claim robustness to all future laundering or obfuscation strategies.

Finally, broad deployment of AI-generated text detectors may have downstream social effects, including over-policing legitimate writing assistance or disadvantaging users whose writing style differs from the training distribution. For these reasons, such systems should be deployed with clear uncertainty communication, human oversight, and regular re-evaluation on new domains and attack settings.

\section{Inference Latency and Compute Overhead}
\label{appendix:latency}

All measurements were obtained on a single NVIDIA A100-PCIe GPU (40\,GB) in \texttt{float16}. We benchmark inference on 64 texts from MIRAGE-DIG/\textsc{Generate} (mean length: 980 characters), truncated to 512 tokens. Each detector is warmed up for three runs before timing. We report per-text latency at batch size \(1\), throughput at batch size \(16\), total parameter count (frozen backbone plus detector head), and peak GPU memory. We will release code to reproduce all experiments upon acceptance.

\begin{table*}[t]
\centering
\small
\setlength{\tabcolsep}{6pt}
\begin{tabular}{l r r r r}
\toprule
\textbf{Detector} & \textbf{Params} & \textbf{Latency} & \textbf{Throughput} & \textbf{Peak GPU} \\
& & \textbf{(ms/text, \(b{=}1\))} & \textbf{(texts/s, \(b{=}16\))} & \textbf{(MB)} \\
\midrule
SV-Detect (ours) & 2.65\,B & 25.71 & 74.3  & 8,951 \\
Log-Likelihood   & 2.65\,B & 27.63 & 75.1  & 7,746 \\
Fast-DetectGPT   & 2.65\,B & 49.79 & 37.2  & 10,703 \\
\midrule
RoBERTa-Base     & 0.12\,B & 7.21  & 1,295 & 327 \\
RoBERTa-Large    & 0.36\,B & 11.40 & 714   & 783 \\
\bottomrule
\end{tabular}
\caption{Inference cost comparison on an A100 40\,GB (\texttt{fp16}, 512-token cap, MIRAGE-DIG texts). Parameter counts include the frozen backbone and the detector head. For SV-Detect, the detector-specific overhead is only a small logistic-regression head on top of the LM representations.}
\label{tab:latency}
\end{table*}

\paragraph{SV-Detect adds almost no overhead beyond a single LM forward pass.}
SV-Detect runs in \(25.7\) ms per text, essentially matching a bare Log-Likelihood scoring pass through the same GPT-Neo-2.7B backbone (\(27.6\) ms). The small advantage in favor of SV-Detect is expected: Log-Likelihood requires a full \(50{,}257\)-way log-softmax at every token, whereas SV-Detect only uses hidden states, followed by lightweight mean pooling, cosine projections, and a tiny logistic-regression head. In practice, the detector-specific computation is negligible compared to the backbone forward pass.

\paragraph{SV-Detect is much cheaper than perturbation-based LLM detectors.}
Fast-DetectGPT requires two forward passes through a 2.7B-class model and is therefore about \(1.9\times\) slower than SV-Detect in our setup (\(49.8\) ms vs.\ \(25.7\) ms per text). The gap would be even larger for DetectGPT, whose cost scales linearly with the number of perturbations. Using the standard \(K{=}100\) setting, a simple extrapolation from the Log-Likelihood baseline yields roughly \(2.8\) s per text, i.e.\ around \(100\times\) the cost of SV-Detect. We therefore do not benchmark DetectGPT separately, since its overhead is determined directly by repeated LM forward passes.

\paragraph{RoBERTa-based supervised detectors are cheaper, but less robust.}
RoBERTa-Base and RoBERTa-Large are substantially faster than SV-Detect, with latencies of \(7.2\) ms and \(11.4\) ms per text, respectively. The throughput difference is even larger at batch size \(16\), reflecting both smaller parameter counts and lower memory pressure. This is the main compute--robustness trade-off of our approach: SV-Detect inherits the cost of an LLM forward pass, whereas small encoder-based detectors are far cheaper. However, as shown in the main experiments, these lighter supervised baselines degrade substantially under source-model and attack-family shifts. In contrast, SV-Detect remains close to the cost of a single zero-shot LM forward while providing much stronger generalization.

\paragraph{Training overhead is modest.}
The training cost of SV-Detect is dominated by a single offline pass through the reference LM to extract layer-wise activations, which can then be cached once per benchmark. The downstream classifier operates on only \(L\) or \(L \times 3\) features and fits in seconds on a CPU. By contrast, RoBERTa-based supervised baselines require full encoder fine-tuning for each training setup, which is the main difference in training cost rather than inference cost.
\clearpage

\section{Ablation on the choice of LLM backbone}
\label{sec:ablation_backbone}

In the main experiments, SV-Detect uses GPT-Neo as the probe LLM in order to match the backbone used by prior work and enable a fair comparison. Here, we study how the method changes when the probe backbone is replaced with a different small open-weight LLM. We consider four alternatives with parameter counts broadly comparable to GPT-Neo: \texttt{Qwen/Qwen3-1.7B}, \texttt{Qwen/Qwen3-1.7B-Base}, \texttt{google/gemma-3-1b-pt}, and \texttt{google/gemma-3-1b-it}. This also lets us compare pretrained/base and instruction-tuned variants within the same model family.

Table~\ref{tab:ablation_on_llm} reports AUROC on DetectRL Multi-Domain transfer. Overall, all tested backbones perform strongly, with near-perfect diagonal performance and robust off-diagonal transfer. The strongest results are obtained with the Qwen backbones. Both Qwen3-1.7B and Qwen3-1.7B-Base improve over GPT-Neo on most train--test pairs, and Qwen3-1.7B-Base is the strongest overall. In particular, it gives the best off-diagonal transfer when trained on \textsc{ArXiv}, \textsc{XSum}, \textsc{Writing}, or \textsc{Review}, suggesting that its representation geometry is especially well suited to learning stable human-vs-AI directions.

GPT-Neo remains competitive, but is generally weaker than the Qwen variants on the harder cross-domain pairs, especially those involving transfer into or out of \textsc{Writing}. For example, the \textsc{ArXiv} \(\rightarrow\) \textsc{Writing} and \textsc{XSum} \(\rightarrow\) \textsc{Writing} pairs improve substantially under both Qwen backbones.

The Gemma backbones are also competitive. Gemma3-1B-pt performs strongly across most transfer pairs and often approaches GPT-Neo, while Gemma3-1B-it achieves similarly high diagonal performance and remains robust under transfer. Overall, Gemma backbones remain slightly less consistent than Qwen on the hardest off-diagonal pairs, but still yield strong AUROC throughout.

Taken together, these results indicate that SV-Detect is not tied to GPT-Neo and remains effective across several backbone families. Backbone choice still matters, however: Qwen-style representations provide the strongest and most stable transfer, while GPT-Neo and Gemma remain competitive but slightly less robust on the hardest cross-domain pairs.
\begin{table}[t]
  \centering
  \scriptsize
  \setlength{\tabcolsep}{4pt}
  \renewcommand{\arraystretch}{1.02}

  \begin{tabular}{lcccc}
  \toprule
  \multicolumn{5}{c}{\textbf{GPT-Neo-2.7B}} \\
  \midrule
  Train & ArX. & XSum & Writ. & Rev. \\
  \midrule
  ArXiv   & 100.00 &  97.34 &  86.82 &  94.05 \\
  XSum    &  97.80 &  99.94 &  83.02 &  95.24 \\
  Writing &  86.74 &  81.82 &  99.98 &  99.40 \\
  Review  &  94.17 &  93.96 &  99.38 &  99.96 \\
  \midrule
  \multicolumn{5}{c}{\textbf{Qwen3-1.7B}} \\
  \midrule
  Train & ArX. & XSum & Writ. & Rev. \\
  \midrule
  ArXiv   & 100.00 &  99.34 &  87.49 &  96.28 \\
  XSum    &  99.40 &  99.98 &  96.29 &  98.84 \\
  Writing &  91.23 &  91.69 &  99.98 &  99.60 \\
  Review  &  97.24 &  98.43 &  99.48 &  99.93 \\
  \midrule
  \multicolumn{5}{c}{\textbf{Qwen3-1.7B-Base}} \\
  \midrule
  Train & ArX. & XSum & Writ. & Rev. \\
  \midrule
  ArXiv   & 100.00 &  99.24 &  94.21 &  98.19 \\
  XSum    &  99.50 &  99.98 &  97.39 &  99.41 \\
  Writing &  95.29 &  94.02 &  99.99 &  99.76 \\
  Review  &  97.87 &  97.94 &  99.77 &  99.96 \\
  \midrule
  \multicolumn{5}{c}{\textbf{Gemma3-1B-pt}} \\
  \midrule
  Train & ArX. & XSum & Writ. & Rev. \\
  \midrule
  ArXiv   & 100.00 &  98.26 &  89.97 &  96.51 \\
  XSum    &  98.77 &  99.98 &  94.96 &  98.48 \\
  Writing &  86.19 &  93.72 &  99.98 &  99.67 \\
  Review  &  93.72 &  96.41 &  99.32 &  99.97 \\
  \midrule
  \multicolumn{5}{c}{\textbf{Gemma3-1B-it}} \\
  \midrule
  Train & ArX. & XSum & Writ. & Rev. \\
  \midrule
  ArXiv   & 100.00 &  98.83 &  86.38 &  96.05 \\
  XSum    &  98.43 &  99.94 &  95.13 &  98.16 \\
  Writing &  85.76 &  95.58 &  99.98 &  99.30 \\
  Review  &  95.44 &  96.60 &  99.03 &  99.94 \\
  \bottomrule
  \end{tabular}

  \caption{Ablation on the probe LLM backbone for DetectRL Multi-Domain
  transfer, per-cell AUROC (rows = training domain, columns = evaluation
  domain).}
  \label{tab:ablation_on_llm}
  \end{table}

\clearpage

\section{Interpretability Analysis}
\label{sec:interpretation_suppl}

This section provides the full methodology underlying the interpretability analyses summarized in Section~\ref{sec:interpretation}. Specifically, we study: (i) per-layer attribution of the detector, (ii) logit-lens decoding of steering vectors, (iii) a hand-crafted stylistic-feature baseline, and (iv) token-level visualization of where the detection signal concentrates within a text. Unless stated otherwise, all analyses use the same frozen reference model (GPT-Neo-2.7B) and the same trained detectors as in the main experiments.

\subsection{Per-layer contribution to the detector}
\label{sec:layer_contrib}

\paragraph{Setup.}
Our detector is a logistic regression applied to layer-wise projection scores. For DetectRL, we use one feature per transformer layer:
\[
\mathbf{s}(x) = (s_1(x), \dots, s_L(x)) \in \mathbb{R}^L,
\]
\[
s_\ell(x) = \frac{\langle a_\ell(x), v_\ell \rangle}{\|a_\ell(x)\|_2},
\]
where \(a_\ell(x)\) is the mean-pooled residual representation at layer \(\ell\), and \(v_\ell\) is the corresponding steering vector. For MIRAGE, we retain three orthonormalized directions per layer (one per task), so \(\mathbf{s}(x) \in \mathbb{R}^{L \times 3}\) is flattened before being passed to the classifier. In all cases, we train a standardized logistic regression (\texttt{StandardScaler} followed by \(\ell_2\)-regularized \texttt{LogisticRegression} with \texttt{liblinear} solver).

\paragraph{Attribution.}
After fitting the classifier, we attribute importance to each scalar feature using the magnitude-weighted coefficient
\[
c_\ell = |w_\ell| \cdot \sigma_\ell,
\]
where \(w_\ell\) is the learned coefficient and \(\sigma_\ell\) is the empirical standard deviation of feature \(\ell\) on the training set. This quantity is invariant to feature rescaling and corresponds to the effective contribution of a standardized feature to the classifier logit. For MIRAGE, we sum \(c_\ell\) across the three task-specific directions to obtain a single per-layer attribution score.

\begin{figure}[t]
    \centering
    \includegraphics[width=\columnwidth]{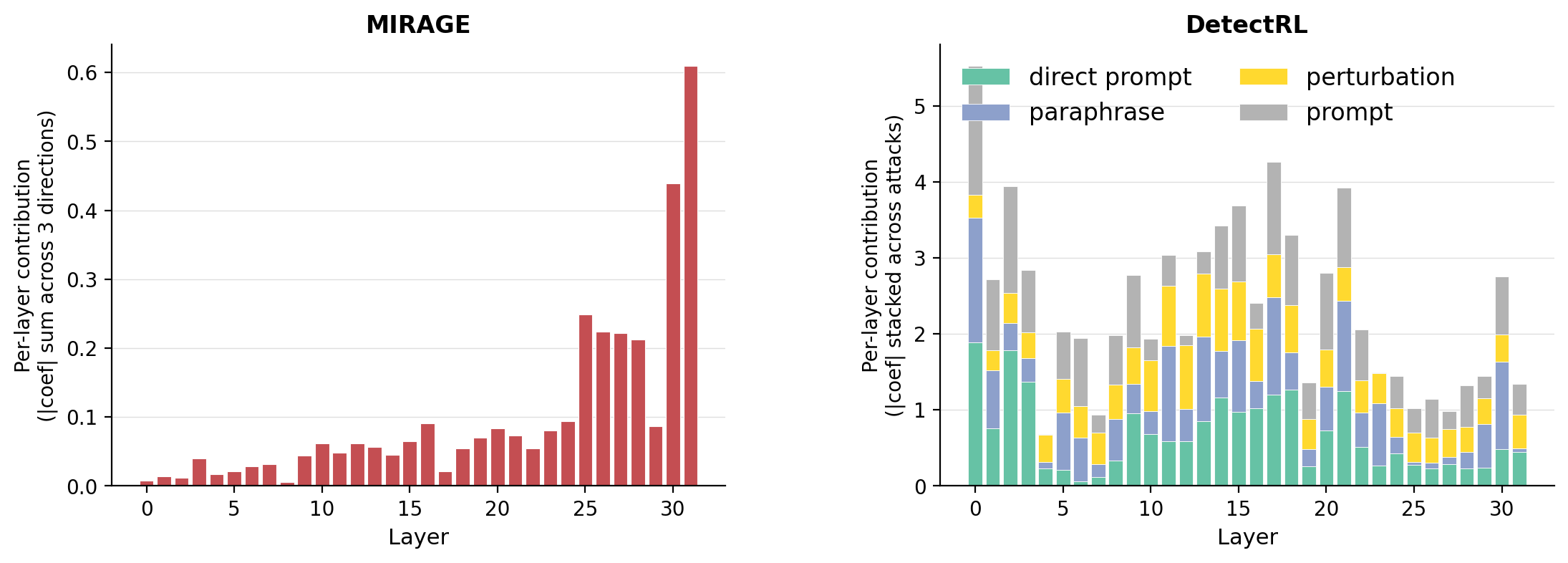}
    \caption{}
    \label{fig:layer_contrib}
\end{figure}

Fig.~\ref{fig:layer_contrib} shows that, on MIRAGE, the detector’s signal is concentrated in the final block of layers (L29-31), consistent with prior observations that higher-level stylistic features tend to emerge closer to the model output. On DetectRL, the contribution is more distributed: there is a sharp peak at L0-2, which we associate with surface-level lexical cues, a broader ridge at L14-18, and a secondary peak around L21. This profile motivates our choice of L25--31 for MIRAGE and L14-18 for DetectRL as the main targets of the logit-lens analysis below.

\subsection{Logit-lens decoding of the steering vectors}
\label{sec:logit_lens}

\paragraph{Method.}
We apply a standard logit-lens probe \citep{nostalgebraist2020} to each layer-\(\ell\) steering vector \(v_\ell\) by projecting it through the model’s final layer norm and unembedding head:
\[
\mathrm{logits}_\ell = W_U^{\top}\,\mathrm{LN}_f(v_\ell)
\in \mathbb{R}^{|\mathcal{V}|},
\]
where \(|\mathcal{V}| = 50{,}257\) is the GPT-Neo-2.7B vocabulary size. The interpretation is directional: the top-ranked indices correspond to tokens aligned with the \(+v_\ell\) direction, which we interpret as \emph{LLM-leaning}, while the bottom-ranked indices correspond to tokens aligned with \(-v_\ell\), i.e.\ the \emph{human-leaning} direction. We report raw logits rather than softmax probabilities, since the goal is to inspect relative alignment with the direction rather than recover a calibrated next-token distribution.


\paragraph{Results.}
Fig.~\ref{fig:logit_lens_tokens} shows the top six tokens on each side. The resulting patterns are consistent across the selected layers:
\begin{itemize}
    \item \textbf{MIRAGE, \(-v_\ell\) (human-leaning).} Stylized closing punctuation (\texttt{!\textquotedbl}, \texttt{...\textquotedbl}, \texttt{').'}), formal sentence-ending subwords (\texttt{authorised}, \texttt{itially}), and decorative or exaggerated tokens (\texttt{\textsterling}, \texttt{?????}).
    \item \textbf{MIRAGE, \(+v_\ell\) (LLM-leaning).} Surnames and journal abbreviations from the arXiv-derived split together with fragments more characteristic of academic prose (\texttt{baugh}, \texttt{foreseen}, \texttt{paralleled}, \texttt{flavorful}).
    \item \textbf{DetectRL, \(+v_\ell\) (LLM-leaning).} Subword fragments of formal or emphatic vocabulary (\texttt{astical}, \texttt{TRY}, \texttt{ermanent}, \texttt{ebted}) and atypical punctuation or Unicode glyphs.
    \item \textbf{DetectRL, \(-v_\ell\) (human-leaning).} More colloquial or topical fragments, including \texttt{organising}, \texttt{Playstation}, \texttt{traged}, \texttt{adolesc}, and \texttt{bigot}.
\end{itemize}

We stress that token-level logit-lens readings are noisy at any single layer. The patterns above should therefore be read as the consensus across the top contributing layers, rather than as literal interpretations of isolated tokens. Tokens that appear only at one layer are likely artifacts and should not be over-interpreted.

\subsection{Stylistic-feature baseline}
\label{sec:regex}

To quantify how much of the detector’s signal can be explained by hand-recognizable surface cues, we train a baseline logistic regression on a fixed bank of regex-derived features and compare its in-distribution AUROC to the full steering vector pipeline.

\paragraph{Feature bank.}
For each text, we compute counts and per-1{,}000-character rates for the following classes of patterns:
\begin{itemize}
    \item \textbf{Typography:} em-dash usage (\texttt{—}), curly quotes, Unicode ellipsis (\texttt{\textellipsis}) versus three-dot ellipsis (\texttt{...}), and space-before-period.
    \item \textbf{Markdown / formatting:} bold, italics, inline code, fenced code blocks, headers, and Oxford-comma lists.
    \item \textbf{Discourse:} formal connectives (\texttt{Moreover}, \texttt{Furthermore}, \texttt{Additionally}, \texttt{In conclusion}, \texttt{Overall}), summary prefixes, casual hedges (\texttt{kinda}, \texttt{sorta}, \texttt{tbh}), and inclusive pronouns (\texttt{we}, \texttt{us}, \texttt{our}).
    \item \textbf{Lexical:} “polished paraphrase” phrasing (e.g.\ \texttt{utilizing}, \texttt{endeavor}, \texttt{showcase}), consonant-cluster words, and log character length.
    \item \textbf{Prompt leakage and template artifacts:} \texttt{[Assistant]:} prefixes, templated review phrasing, generic summary formulas, and refusal-style stubs.
    \item \textbf{Mid-word case flips:} regex patterns that detect uppercase letters inside otherwise lowercase words, a known artifact under aggressive perturbation attacks.
\end{itemize}
We standardize the resulting feature vectors and fit an \(\ell_2\)-regularized logistic regression, tuning the regularization strength \(C\) by 5-fold cross-validation on a held-out split.

\paragraph{Results.}
Table~\ref{tab:regex_metrics} reports the in-distribution AUROC of this stylistic-feature baseline across all evaluated settings, together with the full steering vector pipeline for reference; the corresponding dumbbell visualization is shown in Fig.~\ref{fig:interp_comparison}. Two observations stand out:
\begin{enumerate}
    \item Stylistic features alone form a non-trivial baseline, reaching 76--82 AUROC on MIRAGE and 88--91 AUROC on DetectRL. This is consistent with prior work showing that surface-level statistical and stylistic cues can already carry substantial signal for distinguishing machine-generated from human-written text.
    \item The full steering vector pipeline still improves on this baseline by 13--24 AUROC points on MIRAGE and 9--12 points on DetectRL. We interpret this remaining gap as the part of the signal that is present in hidden representations but not recoverable from the regex bank. At the same time, the highest-weight regex features overlap substantially with the tokens surfaced by the logit-lens analysis (Fig.~\ref{fig:interpretation_combined}), including em-dashes, formal connectives, and prompt-leakage templates.
\end{enumerate}

\subsection{Cross-LM Logit-lens decoding of steering vectors}
\label{sec:cross_lm_logit_lens}

The main interpretability analyses in
Sec~\ref{sec:layer_contrib}
decode steering vectors through the same reference LM that produced them, namely GPT-Neo-2.7B. A natural question is whether the human-vs-AI directions learned by SV-Detect are specific to GPT-Neo's residual stream, or whether similar directions emerge in other modern LMs. To address this, we retrain the steering vector pipeline independently on several additional reference LMs and then decode the resulting dense steering vectors through each model's
own unembedding.

\paragraph{Reference LMs and datasets.}
We use four reference LMs spanning three families: \texttt{EleutherAI/gpt-neo-2.7B}, \texttt{Qwen/Qwen3-1.7B}, \texttt{google/gemma-3-1b-pt}, and \texttt{google/gemma-3-1b-it}. For each LM, we extract per-layer mean-pooled residual activations on the corresponding in-domain training corpora and fit the standard \texttt{logreg} steering vector construction described in Sec.~\ref{sec:methodology}. This yields one dense steering vector per layer for each (LM, dataset)
pair. GPT-Neo is analyzed on four datasets
(MIRAGE \textsc{Generate}/\textsc{Polish}/\textsc{Rewrite} and
DetectRL \texttt{direct\_prompt}), while the other three LMs are
analyzed on these four settings plus four additional DetectRL
per-domain subsets (\texttt{arxiv}, \texttt{xsum},
\texttt{writing\_prompt}, \texttt{yelp\_review}), for a total of
28 (LM, dataset) pairs.

\paragraph{Method.}
For each (LM, dataset, layer), we decode the steering vector
through the LM's own final layer norm and unembedding using logit-lens probe as described in Sec.~\ref{sec:logit_lens}.

Because single-layer readings are noisy and strongly affected by
tokenizer idiosyncrasies, we aggregate within each LM before
comparing across LMs. Specifically, for each LM we pool the
top-$K_{\mathrm{pool}}=8$ tokens at every layer in the last third of the network depth, separately by sign, then take the union across the LM's datasets. Layer ranges are
L17--25 for Gemma-3, L18--27 for Qwen3, and L21--31 for GPT-Neo.
Before computing cross-LM overlap, we normalize tokens by
stripping whitespace, lowercasing, applying NFKC normalization,
dropping control/format characters, and filtering out pure
punctuation as well as tokens with less than 50\% ASCII letters.
This removes tokenizer-specific artifacts and makes the consensus comparison more meaningful.

\paragraph{Results.}
\begin{figure}[t]
    \centering
    \includegraphics[width=\columnwidth]{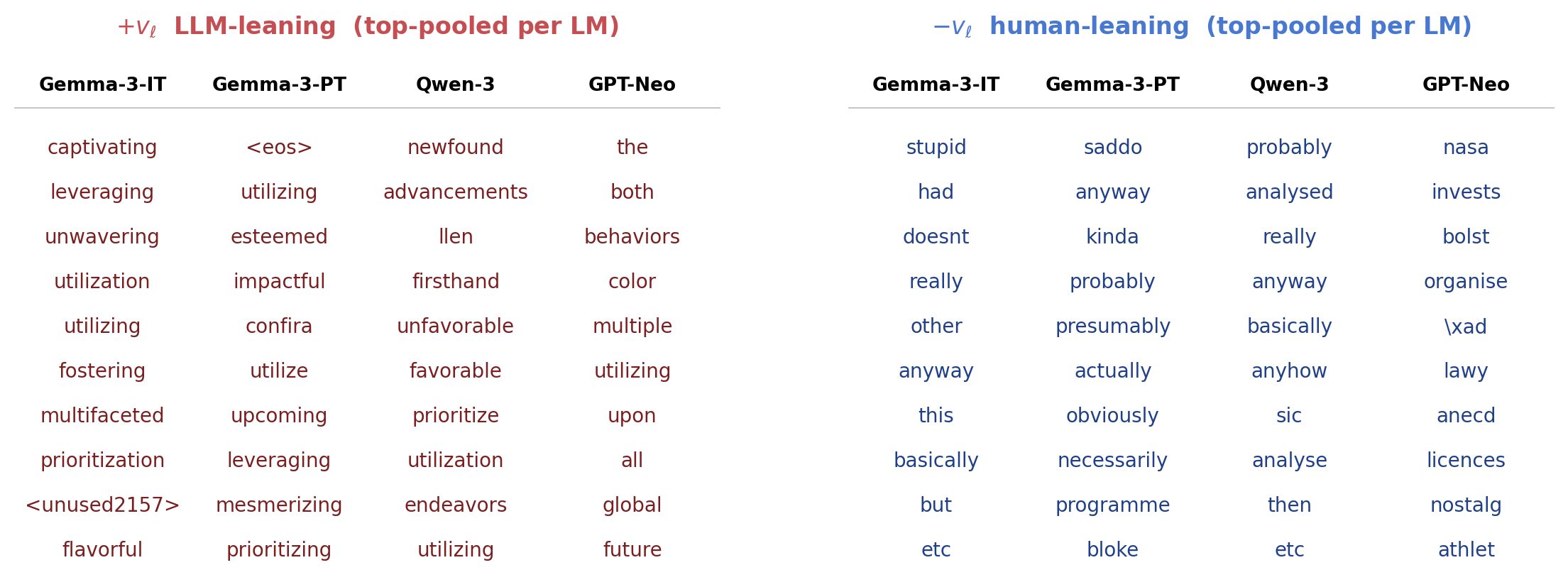}
    \caption{Top-pooled tokens per reference LM, separately for the \(+v_\ell\) (LLM-leaning, left) and
    \(-v_\ell\) (human-leaning, right) sides. For each LM, we pool the top-\(K\) tokens across the last third of layers and across all analyzed DetectRL / MIRAGE settings, then list the most frequent
    normalized tokens. Despite differences in family and tokenizer, the positive side consistently surfaces a polished LLM-style register, while the negative side is more casual and discourse-like.}
    \label{fig:cross_lm_top_per_lm}
\end{figure}


Fig.~\ref{fig:cross_lm_top_per_lm} shows the top pooled tokens
for each LM separately, and
Fig.~\ref{fig:cross_lm_consensus} summarizes the cross-LM
consensus. On the positive side, the four LMs converge on a clear LLM-polished register. Three tokens (
\emph{endeavors}, \emph{utilization}, and
\emph{utilizing}) appear in the top pool of all four LMs, and a
broader cluster including \emph{captivating},
\emph{cybersecurity}, \emph{leveraging},
\emph{optimization}, \emph{favorable}, and
\emph{unfavorable} appears in three of the four. On the negative
side, the consensus is weaker but still interpretable: the most
stable tokens are casual discourse markers such as
\emph{anyway}, \emph{basically}, \emph{etc},
\emph{maybe}, \emph{really}, and \emph{very}. Here, the three
smaller modern LMs agree more strongly with one another, while
GPT-Neo more often surfaces named-entity fragments and topical
content.

Overall, this analysis suggests that the lexical contrast captured by SV-Detect is not specific to GPT-Neo. When the steering vector pipeline is retrained independently on different LM families and their dense directions are decoded through each model's own unembedding, the same broad semantic register reappears: LLM-polished, formal, and abstract language on the positive side, and more casual, discourse-like language on the negative side.

\subsection{Per-token visualization}
\label{appendix:token_viz}

The token-level figure in the main paper (Fig.~\ref{fig:teaser}) is constructed as follows. Given a text \(x\) and the best-AUROC layer $\ell^\star$ identified by an AUROC-probe on the held-out set, we forward \(x\) through the frozen reference model and record the per-token residual vector
\[
h_{\ell^\star}(x_t) \in \mathbb{R}^d
\qquad \text{for } t \in \{1, \dots, T\}.
\]
Each token is then assigned the dot product
\[
\rho_t = \langle h_{\ell^\star}(x_t),\, v_{\ell^\star} \rangle
\]
where \(v_{\ell^\star}\) is the steering vector at layer \(\ell^\star\). Tokens are colored according to \(\rho_t\), with the scale normalized to the 70th percentile of \(|\rho|\) in the corpus so that most informative tokens reach full saturation. Red denotes positive projection (\(+v\), LLM-leaning) and blue denotes negative projection (\(-v\), human-leaning). The classification banner above each text corresponds to the full text-level prediction produced by the trained detector used in the main experiments.


\begin{table}[t]
\centering
\scriptsize
\setlength{\tabcolsep}{3pt}
\renewcommand{\arraystretch}{1.0}
\begin{tabular}{llccc}
\toprule
\textbf{Setting} & \textbf{Subset} & \textbf{Stylistic} & \textbf{SV-Detect} & \textbf{Gap} \\
& & \textbf{AUROC} & \textbf{AUROC} & \\
\midrule
\multicolumn{5}{c}{\textit{MIRAGE-DIG}} \\
\midrule
DIG      & Generate & 76.98 & 99.12 & +22.1 \\
DIG      & Polish   & 80.74 & 95.09 & +14.3 \\
DIG      & Rewrite  & 75.94 & 95.03 & +19.1 \\
\midrule
\multicolumn{5}{c}{\textit{MIRAGE-SIG}} \\
\midrule
SIG      & Generate & 75.59 & 99.09 & +23.5 \\
SIG      & Polish   & 81.71 & 94.99 & +13.3 \\
SIG      & Rewrite  & 76.36 & 94.21 & +17.8 \\
\midrule
\multicolumn{5}{c}{\textit{DetectRL (Multi-Attack)}} \\
\midrule
DetectRL & Direct   & 90.73 & 99.99 & +9.3  \\
DetectRL & Prompt   & 90.11 & 99.96 & +9.9  \\
DetectRL & Paraph.  & 88.69 & 99.98 & +11.3 \\
DetectRL & Perturb. & 87.81 & 99.99 & +12.2 \\
\bottomrule
\end{tabular}
\caption{In-distribution AUROC of the stylistic-feature baseline (Section~\ref{sec:regex}) compared to the full steering vector pipeline. The \textbf{Gap} column shows the additional discriminative signal captured by hidden representations beyond the regex feature bank.}
\label{tab:regex_metrics}
\end{table}


\section{Results on the COLING-2025 MGT benchmark}
\label{sec:coling}

For completeness, we additionally evaluate SV-Detect on the
COLING-2025 Multilingual MGT benchmark, English split
\citep{DBLP:journals/corr/abs-2501-11012}. Unlike DetectRL and MIRAGE, this benchmark pools many generators into a single binary detection task and provides an official held-out test set with labels. It therefore offers a complementary large-scale setting for evaluating the same steering vector pipeline and testing its sensitivity to the composition of the generator pool.

\subsection{Setup}

\paragraph{Data.}
The COLING-2025 MGT English split contains three partitions:
(i) \textbf{train}: 610{,}767 examples (228{,}922 human, 381{,}845 machine-generated from 27 generators),
(ii) \textbf{dev}: 261{,}758 examples (98{,}328 human, 163{,}430 machine-generated), and
(iii) \textbf{test}: 73{,}941 examples, released with ground-truth labels for leaderboard evaluation.
The task is binary classification (human vs.\ machine-generated).

\paragraph{Backbones.}
We evaluate SV-Detect with two reference LMs: the
\texttt{EleutherAI/gpt-neo-2.7B} used in the rest of the paper, and the larger \texttt{meta-llama/Llama-2-7b-hf}. For each backbone, we extract per-layer mean-pooled residual activations and use the same downstream logistic-regression detector as in the main paper.

\paragraph{Ablation axes.}
We vary two components of the pipeline:
\begin{enumerate}
    \item \textbf{Steering vector construction.}
    We compare the three constructions used elsewhere in the paper:
    \emph{Mean} (normalized class-mean difference),
    \emph{PCA} (leading principal component of AI-minus-human activation differences)
    and \emph{LogReg} (\(\ell_2\)-regularized logistic regression on activations, our default construction).
    \item \textbf{Generator-pool filter.}
    The training split contains a long tail of weak open-source generators whose outputs are often much less realistic than those of stronger LLMs. To test whether these generators help or hurt, we compare:
    \begin{itemize}
        \item \emph{all}: all 27 generators
        \item \emph{trimmed}: remove 20 weak generators (Dolly, BLOOMZ, OPT-*, Flan-T5-*, T0-*), dropping 133{,}193 samples and retaining 248{,}652 AI-generated examples. Generators are only removed from the training and dev splits, the whole test split is still used.
    \end{itemize}
\end{enumerate}

We use the same logistic regression detector as described in Sec.~\ref{sec:methodology}. The downstream classifier is  used elsewhere in the paper. We sweep \(C \in \{10^{-3},10^{-2},10^{-1},1,10,100\}\) and select the best value on dev AUROC. All other hyperparameters match the main experiments as described in Sec.~\ref{sec:experiments_setup}.

\subsection{Results}

Llama-based SV-Detect outperforms all baselines by a substantial margin. Removing weak models from the training data helps Llama-based SV-Detect further, gaining about 2 points in Accuracy and 1.7 in $F_1$. However, for GPT-Neo-based SV-Detect, removing weak models hurts. We attribute this to the size and capability of the LM backbone: larger and more capable backbones, such as Llama, benefit from a cleaner and more homogeneous generator pool, whereas smaller backbones like GPT-Neo appear to benefit from the additional diversity provided by weaker generators.

\begin{table}[t]
\centering
\scriptsize
\setlength{\tabcolsep}{3pt}
\renewcommand{\arraystretch}{0.95}
\resizebox{\columnwidth}{!}{%
\begin{tabular}{r l c c}
\toprule
\textbf{Rank} & \textbf{Team / Method} & \textbf{Macro-F1} & \textbf{Acc.} \\
\midrule
1  & \textbf{SV-Detect (Llama-2-7b, trimmed)}            & \textbf{84.8} & \textbf{85.00} \\
2  & \underline{SV-Detect (Llama-2-7b, all)}            & \textbf{83.7} & \textbf{84.00} \\
3 & Advacheck                         & 83.07 & 83.11 \\
4  & Unibuc-NLP                        & 83.01 & 83.33 \\
5  & Fraunhofer SIT                    & 82.80 & 82.89 \\
6  & Grape                             & 81.88 & 82.23 \\
7  & TechExperts(IPN)                  & 81.53 & 81.81 \\
8  & TurQUaz                           & 80.68 & 80.74 \\
9  & SzegedAI                          & 79.10 & 79.29 \\
10  & AAIG                              & 78.74 & 79.34 \\
11  & DCBU                              & 77.13 & 78.01 \\
12  & \underline{SV-Detect (GPT-Neo, all)}            & \textbf{75.6} & \textbf{76.50} \\
13 & Alfa                              & 75.37 & 76.42 \\
14 & L3i++                             & 74.63 & 75.54 \\
15 & LuxVeri                           & 74.58 & 75.68 \\
16 & azlearning                        & 74.14 & 75.17 \\
17  & \underline{SV-Detect (GPT-Neo, trimmed)}            & \textbf{74.0} & \textbf{75.50} \\
18 & honghanhh                         & 73.94 & 75.14 \\
-- & Baseline                          & 73.42 & 74.89 \\
19 & VX1291                            & 72.93 & 74.83 \\
20 & cuettransform                     & 72.32 & 73.16 \\
21 & rockstart                         & 72.24 & 73.89 \\
22 & batirsdu                          & 71.01 & 71.42 \\
23 & IPN-CIC                           & 70.68 & 72.42 \\
24 & Ai-Monitors                       & 70.57 & 72.65 \\
25 & semanticcuet                      & 70.05 & 71.96 \\
26 & hmcgovern                         & 68.48 & 69.51 \\
27 & abhirak0603                       & 68.02 & 70.50 \\
28 & cnlpnitspp                        & 65.02 & 68.76 \\
29 & mail6djj                          & 64.66 & 68.46 \\
30 & bennben                           & 63.32 & 67.48 \\
31 & saehyunma                         & 62.80 & 67.25 \\
32 & yuwert777                         & 62.14 & 66.69 \\
33 & seven                             & 59.09 & 63.20 \\
34 & fangsifan                         & 58.48 & 62.68 \\
35 & yaoxy                             & 57.28 & 64.20 \\
36 & jojoc                             & 54.16 & 60.37 \\
37 & dominikmacko                      & 49.94 & 50.78 \\
38 & tropaleum                         & 49.57 & 50.60 \\
39 & starlight1                        & 47.57 & 56.65 \\
40 & nitstejasrikar                    & 44.89 & 57.24 \\
\bottomrule
\end{tabular}%
}
\caption{Adapted English leaderboard from Table~4 of \citet{DBLP:journals/corr/abs-2501-11012}, with our four LogReg-based SV-Detect variants inserted. Ranking is by Macro-F1.}
\label{tab:coling}
\end{table}

\subsection{Ablation on steering vector construction}
Tab.~\ref{tab:coling_ablation_gptneo} and Tab.~\ref{tab:coling_ablation_llama} report results for all six combinations of steering vector construction and generator-pool filter on both backbones. As in DetectRL and MIRAGE, the LogReg construction consistently outperforms Mean and PCA. The larger Llama-2-7B backbone further improves substantially over GPT-Neo-2.7B.

\begin{table}[t]
\centering
\scriptsize
\setlength{\tabcolsep}{4pt}
\renewcommand{\arraystretch}{0.98}
\begin{tabular}{l l c c c c}
\toprule
\multicolumn{6}{c}{\textbf{GPT-Neo-2.7B}} \\
\midrule
\textbf{SV} & \textbf{Pool} & \textbf{Dev} & \textbf{Test} & \textbf{Acc.} & \textbf{\(F_1\)} \\
\textbf{construction} & & \textbf{AUROC} & \textbf{AUROC} & & \\
\midrule
Mean   & all     & 0.865 & 0.860 & 0.718 & 0.696 \\
Mean   & trimmed & 0.855 & 0.856 & 0.737 & 0.721 \\
PCA    & all     & 0.740 & 0.751 & 0.559 & 0.439 \\
PCA    & trimmed & 0.737 & 0.753 & 0.557 & 0.433 \\
LogReg & all     & \textbf{0.975} & 0.866 & \textbf{0.765} & \textbf{0.756} \\
LogReg & trimmed & 0.952 & \textbf{0.873} & 0.755 & 0.740 \\
\bottomrule
\end{tabular}
\caption{SV-Detect on the COLING-2025 English benchmark with a GPT-Neo-2.7B backbone.}
\label{tab:coling_ablation_gptneo}
\end{table}

\begin{table}[t]
\centering
\scriptsize
\setlength{\tabcolsep}{4pt}
\renewcommand{\arraystretch}{0.98}
\begin{tabular}{l l c c c c}
\toprule
\multicolumn{6}{c}{\textbf{Llama-2-7B}} \\
\midrule
\textbf{SV} & \textbf{Pool} & \textbf{Dev} & \textbf{Test} & \textbf{Acc.} & \textbf{\(F_1\)} \\
\textbf{construction} & & \textbf{AUROC} & \textbf{AUROC} & & \\
\midrule
Mean   & all     & 0.922 & 0.912 & 0.716 & 0.690 \\
Mean   & trimmed & 0.908 & 0.887 & 0.694 & 0.666 \\
PCA    & all     & 0.864 & 0.796 & 0.650 & 0.615 \\
PCA    & trimmed & 0.858 & 0.798 & 0.647 & 0.610 \\
LogReg & all     & \textbf{0.984} & \textbf{0.938} & 0.840 & 0.837 \\
LogReg & trimmed & 0.974 & 0.937 & \textbf{0.850} & \textbf{0.848} \\
\bottomrule
\end{tabular}
\caption{SV-Detect on the COLING-2025 English benchmark with a Llama-2-7B backbone.}
\label{tab:coling_ablation_llama}
\end{table}

\clearpage
\section{Results Tables}
\label{sec:thorough_tables}

In this section, we provide comprehensive tables with results for all experiments in Sec.~\ref{sec:experiments}.

\subsection{DetectRL}

Tables~\ref{tab:train_test_detectrl_without_generalization_1_vector} and~\ref{tab:train_test_detectrl_with_generalization_1_vector} report the in-domain and cross-source performance of SV-Detect and the baselines, respectively. Fig.~\ref{fig:detectrl_indomain_auroc}, Fig.~\ref{fig:detectrl_indomain_f1} and Fig.~\ref{fig:detectrl_crosssource} provide a visual summary of these results.

The RepreGuard row of Table~\ref{tab:train_test_detectrl_without_generalization_1_vector} is reproduced from the official implementation using our default GPT-Neo-2.7B backbone. For LRR / DetectGPT / Fast-DetectGPT / SV-Detect, we use the training and evaluation protocol described in Sec.~\ref{sec:experiments_setup}. Results show that RepreGuard with GPT-Neo-2.7B as a backbone collapses to near-chance across almost all cells. Additional comparison of SV-Detect with RepreGuard using LLaMA-3.1-8B as a backbone (default backbone of RepreGuard) is reported in Appendix~\ref{sec:appendix_repreguard_vs_svdetect}.

\begin{table*}[t]
\centering
\small
\setlength{\tabcolsep}{4pt}
\renewcommand{\arraystretch}{1.02}

\begin{tabular}{l c c c c c c c c}
\toprule
& \multicolumn{2}{c}{ArXiv} & \multicolumn{2}{c}{XSum} & \multicolumn{2}{c}{Writing} & \multicolumn{2}{c}{Review} \\
\cmidrule(lr){2-3}\cmidrule(lr){4-5}\cmidrule(lr){6-7}\cmidrule(lr){8-9}
Method & AUROC & $F_1$ & AUROC & $F_1$ & AUROC & $F_1$ & AUROC & $F_1$ \\
\midrule
\multicolumn{9}{c}{\textbf{Multi-Domain}}\\
\midrule
Log-Likelihood    & 65.35 & 57.55 & 45.68 & 41.32 & 68.00 & 59.38 & 75.84 & 67.22 \\
Entropy           & 48.39 & 29.71 & 67.84 & 57.23 & 39.06 & 20.55 & 28.82 & 2.14 \\
Rank              & 57.17 & 54.62 & 36.87 & 22.47 & 56.26 & 50.90 & 55.08 & 51.90 \\
Log-Rank          & 67.01 & 60.09 & 46.74 & 42.60 & 67.58 & 57.57 & 76.40 & 69.88 \\
LRR               & 70.54 & 61.34 & 50.09 & 38.38 & 64.65 & 53.09 & 76.61 & 68.99 \\
NPR               & 53.85 & 49.65 & 34.59 & 18.31 & 54.96 & 52.30 & 50.09 & 45.39 \\
DetectGPT         & 22.15 & 0.00  & 12.21 & 0.00  & 58.95 & 50.83 & 44.43 & 35.25 \\
Revise-Detect.    & 70.40 & 37.51 & 50.34 & 46.07 & 73.24 & 64.29 & 75.01 & 68.71 \\
Binoculars        & 84.03 & 76.77 & 77.39 & 72.18 & 94.38 & 79.73 & 90.00 & 84.32 \\
Fast-DetectGPT    & 43.69 & 24.46 & 39.19 & 28.39 & 74.21 & 67.84 & 77.02 & 71.62 \\
RepreGuard\textsuperscript{\dag} & 50.00 & 0.00 & 50.00 & 0.00 & 60.25 & 60.11 & 55.15 & 24.30 \\
\addlinespace[1pt]
RoBERTa-Base      & 100.0 & 100.0 & 99.99 & 99.85 & 99.99 & 99.65 & 99.97 & 99.50 \\
RoBERTa-Large     & 99.99 & 99.90 & 99.85 & 98.95 & 99.54 & 97.73 & 99.76 & 98.90 \\
XLM-RoBERTa-Base  & 100.0 & 100.0 & 99.97 & 99.55 & 99.84 & 98.76 & 99.88 & 99.05 \\
XLM-RoBERTa-Large & 99.98 & 99.85 & 99.84 & 98.95 & 99.85 & 98.31 & 96.40 & 92.66 \\
\addlinespace[1pt]
\textbf{SV-Detect} & \textbf{100.0} & \textbf{100.0} & \textbf{99.94} & \textbf{99.45} & \textbf{99.98} & \textbf{99.60} & \textbf{99.96} & \textbf{99.30} \\
\bottomrule
\end{tabular}

\vspace{1.2ex}

\begin{tabular}{l c c c c c c c c}
\toprule
& \multicolumn{2}{c}{GPT-3.5} & \multicolumn{2}{c}{Claude} & \multicolumn{2}{c}{PaLM-2} & \multicolumn{2}{c}{Llama-2} \\
\cmidrule(lr){2-3}\cmidrule(lr){4-5}\cmidrule(lr){6-7}\cmidrule(lr){8-9}
Method & AUROC & $F_1$ & AUROC & $F_1$ & AUROC & $F_1$ & AUROC & $F_1$ \\
\midrule
\multicolumn{9}{c}{\textbf{Multi-LLM}}\\
\midrule
Log-Likelihood    & 62.89 & 57.80 & 43.32 & 28.10 & 70.03 & 60.73 & 75.65 & 65.90 \\
Entropy           & 46.84 & 23.29 & 52.25 & 30.42 & 45.34 & 16.56 & 43.48 & 66.75 \\
Rank              & 52.19 & 49.32 & 41.68 & 22.78 & 50.40 & 41.74 & 57.05 & 54.40 \\
Log-Rank          & 62.84 & 56.87 & 43.32 & 30.12 & 70.89 & 63.09 & 77.97 & 66.66 \\
LRR               & 61.61 & 52.12 & 43.30 & 18.91 & 71.17 & 65.51 & 83.65 & 75.51 \\
NPR               & 50.29 & 43.81 & 41.64 & 32.91 & 44.64 & 34.77 & 52.53 & 48.68 \\
DetectGPT         & 43.46 & 26.27 & 32.86 & 12.56 & 26.72 & 0.00  & 36.71 & 20.40 \\
DNA-GPT           & 61.87 & 55.04 & 48.88 & 25.67 & 71.48 & 60.77 & 75.22 & 62.89 \\
Revise-Detect.    & 70.10 & 62.72 & 49.87 & 27.28 & 69.84 & 59.03 & 75.65 & 65.87 \\
Binoculars        & 88.14 & 82.50 & 55.15 & 39.35 & 93.30 & 88.20 & 96.64 & 92.30 \\
Fast-DetectGPT    & 65.56 & 59.55 & 30.01 & 0.00  & 65.99 & 57.58 & 76.79 & 69.08 \\
RepreGuard\textsuperscript{\dag} & 66.25 & 67.53 & 50.00 & 0.00 & 50.00 & 0.00 & 50.00 & 0.00 \\
\addlinespace[1pt]
RoBERTa-Base      & 99.97 & 99.70 & 99.98 & 99.80 & 99.94 & 99.40 & 99.84 & 99.45 \\
RoBERTa-Large     & 99.77 & 98.86 & 96.23 & 92.48 & 97.93 & 92.64 & 86.72 & 76.17 \\
XLM-RoBERTa-Base  & 99.88 & 99.45 & 98.26 & 97.48 & 98.77 & 97.19 & 99.69 & 98.57 \\
XLM-RoBERTa-Large & 99.55 & 97.56 & 91.67 & 84.24 & 98.73 & 94.43 & 99.66 & 97.67 \\
\addlinespace[1pt]
\textbf{SV-Detect} & \textbf{99.99} & \textbf{99.60} & \textbf{99.91} & \textbf{98.91} & \textbf{99.88} & \textbf{99.36} & \textbf{99.99} & \textbf{99.60} \\
\bottomrule
\end{tabular}

\vspace{1.2ex}

\begin{tabular}{l c c c c c c c c c c}
\toprule
& \multicolumn{2}{c}{Direct} & \multicolumn{2}{c}{Prompt} & \multicolumn{2}{c}{Paraph.} & \multicolumn{2}{c}{Perturb.} & \multicolumn{2}{c}{Mixing} \\
\cmidrule(lr){2-3}\cmidrule(lr){4-5}\cmidrule(lr){6-7}\cmidrule(lr){8-9}\cmidrule(lr){10-11}
Method & AUROC & $F_1$ & AUROC & $F_1$ & AUROC & $F_1$ & AUROC & $F_1$ & AUROC & $F_1$ \\
\midrule
\multicolumn{11}{c}{\textbf{Multi-Attack}}\\
\midrule
Log-Likelihood    & 89.25 & 82.09 & 86.87 & 78.16 & 64.55 & 57.59 & 35.51 & 0.78 & 63.70 & 53.31 \\
Entropy           & 26.47 & 0.00  & 26.18 & 0.00  & 48.12 & 26.01 & 68.62 & 68.95 & 49.37 & 28.52 \\
Rank              & 83.50 & 76.27 & 81.21 & 72.86 & 60.60 & 52.60 & 8.04  & 0.00 & 52.05 & 42.46 \\
Log-Rank          & 89.25 & 81.45 & 86.35 & 77.51 & 64.69 & 59.17 & 37.71 & 0.78 & 64.63 & 56.86 \\
LRR               & 85.83 & 77.40 & 80.80 & 74.30 & 63.99 & 55.20 & 45.91 & 29.27 & 66.12 & 53.81 \\
NPR               & 77.98 & 71.61 & 77.15 & 70.63 & 56.94 & 46.25 & 6.78  & 0.00 & 48.63 & 37.65 \\
DetectGPT         & 52.84 & 40.90 & 51.83 & 37.98 & 31.79 & 16.89 & 18.21 & 0.00 & 26.28 & 0.00 \\
DNA-GPT           & 88.01 & 80.78 & 85.62 & 77.47 & 65.61 & 54.94 & 40.45 & 2.73 & 62.14 & 50.89 \\
Revise-Detect.    & 86.88 & 79.61 & 84.89 & 76.21 & 67.26 & 62.03 & 43.98 & 7.56 & 65.27 & 54.39 \\
Binoculars        & 94.87 & 89.73 & 93.45 & 88.12 & 88.34 & 81.56 & 76.89 & 69.34 & 89.12 & 83.67 \\
Fast-DetectGPT    & 79.56 & 72.45 & 78.43 & 70.34 & 70.12 & 62.89 & 49.56 & 41.23 & 67.23 & 59.78 \\
RepreGuard\textsuperscript{\dag} & 52.78 & 29.59 & 50.00 & 0.00 & 52.18 & 26.75 & 54.81 & 28.44 & 54.61 & 56.41 \\
\addlinespace[1pt]
RoBERTa-Base      & 99.87 & 99.60 & 99.78 & 99.47 & 99.67 & 99.12 & 98.32 & 97.45 & 99.12 & 98.76 \\
RoBERTa-Large     & 98.73 & 97.83 & 98.45 & 97.56 & 97.89 & 96.78 & 96.12 & 94.67 & 97.56 & 96.34 \\
XLM-RoBERTa-Base  & 99.56 & 99.12 & 99.23 & 99.01 & 98.89 & 98.34 & 98.56 & 97.89 & 99.01 & 98.56 \\
XLM-RoBERTa-Large & 99.45 & 98.67 & 98.89 & 97.98 & 98.23 & 97.67 & 97.89 & 96.34 & 98.67 & 97.89 \\
\addlinespace[1pt]
\textbf{SV-Detect} & \textbf{99.99} & \textbf{99.70} & \textbf{99.96} & \textbf{99.56} & \textbf{99.98} & \textbf{99.55} & \textbf{99.99} & \textbf{99.95} & \textbf{99.83} & \textbf{98.75} \\
\bottomrule
\end{tabular}

\caption{Performance of detectors on DetectRL in the Multi-Domain,
Multi-LLM, and Multi-Attack settings.
\textsuperscript{\dag}RepreGuard is reproduced under the DetectRL training
and evaluation protocol of Sec.~\ref{sec:experiments_setup}, using the
same GPT-Neo-2.7B surrogate as our other rows.}
\label{tab:train_test_detectrl_without_generalization_1_vector}
\end{table*}

\begin{table*}[t]
\centering
\small
\setlength{\tabcolsep}{4pt}
\renewcommand{\arraystretch}{1.03}

\begin{tabular}{l cccc cccc cccc}
\toprule
& \multicolumn{4}{c}{\textbf{SV-Detect}} 
& \multicolumn{4}{c}{\textbf{Fast-DetectGPT}} 
& \multicolumn{4}{c}{\textbf{RoBERTa-Base}} \\
\cmidrule(lr){2-5}\cmidrule(lr){6-9}\cmidrule(lr){10-13}
Train & ArXiv & XSum & Writing & Review & ArXiv & XSum & Writing & Review & ArXiv & XSum & Writing & Review \\
\midrule
\multicolumn{13}{c}{\textbf{Multi-Domain} ($F_1$)}\\
\midrule
ArXiv   & 100.00 & 93.32 & 80.78 & 88.89 & 24.46 & 23.71 & 59.67 & 60.17 & 100.0 & 75.90 & 77.68 & 70.69 \\
XSum    &  93.09 & 99.45 & 75.72 & 88.93 & 28.43 & 28.39 & 62.99 & 63.08 & 68.43 & 99.85 & 71.79 & 67.17 \\
Writing &  80.57 & 75.01 & 99.60 & 96.64 & 34.81 & 33.60 & 67.84 & 68.30 & 78.58 & 72.72 & 99.65 & 94.24 \\
Review  &  87.93 & 87.60 & 97.09 & 99.30 & 40.70 & 37.66 & 68.25 & 71.62 & 82.64 & 84.15 & 85.10 & 99.50 \\
\bottomrule
\end{tabular}

\vspace{1.2ex}

\begin{tabular}{l cccc cccc cccc}
\toprule
& \multicolumn{4}{c}{\textbf{SV-Detect}} 
& \multicolumn{4}{c}{\textbf{Fast-DetectGPT}} 
& \multicolumn{4}{c}{\textbf{RoBERTa-Base}} \\
\cmidrule(lr){2-5}\cmidrule(lr){6-9}\cmidrule(lr){10-13}
Train & GPT-3.5 & PaLM-2 & Claude & Llama-2 & GPT-3.5 & PaLM-2 & Claude & Llama-2 & GPT-3.5 & PaLM-2 & Claude & Llama-2 \\
\midrule
\multicolumn{13}{c}{\textbf{Multi-LLM} ($F_1$)}\\
\midrule
GPT-3.5 & 99.60 & 86.58 & 90.80 & 97.36 & 59.55 & 59.56 & 12.96 & 69.93 & 99.97 & 70.34 & 62.90 & 94.68 \\
PaLM-2  & 99.60 & 98.91 & 94.83 & 99.26 & 55.77 & 57.58 & 8.20  & 68.43 & 99.25 & 99.40 & 93.43 & 99.25 \\
Claude  & 99.45 & 94.87 & 99.36 & 98.21 & 0.19  & 0.00  & 0.00  & 1.18  & 96.83 & 83.92 & 99.80 & 89.77 \\
Llama-2 & 99.55 & 96.41 & 92.13 & 99.60 & 56.28 & 57.74 & 8.65  & 69.08 & 99.45 & 93.02 & 87.56 & 99.45 \\
\bottomrule
\end{tabular}

\vspace{1.2ex}

\begin{tabular}{l cccc cccc cccc}
  \toprule
  & \multicolumn{4}{c}{\textbf{SV-Detect}}
  & \multicolumn{4}{c}{\textbf{Fast-DetectGPT}}
  & \multicolumn{4}{c}{\textbf{RoBERTa-Base}} \\
  \cmidrule(lr){2-5}\cmidrule(lr){6-9}\cmidrule(lr){10-13}
  Train & Prompt & Paraph. & Perturb. & Mixing & Prompt & Paraph. & Perturb. & Mixing & Prompt & Paraph. & Perturb. & Mixing \\
  \midrule
  \multicolumn{13}{c}{\textbf{Multi-Attack} ($F_1$)}\\
  \midrule
  Direct     & 98.93 & 96.54 & 95.46 & 96.91 & 64.01 & 40.45 & 41.02 & 31.81 & 95.73 & 94.91 & 64.32 & 89.07 \\
  Prompt     & 99.56 & 96.00 & 91.66 & 96.45 & 64.00 & 39.94 & 40.40 & 31.25 & 97.18 & 94.98 & 86.18 & 92.92 \\
  Paraphrase & 96.85 & 99.55 & 95.89 & 97.91 & 61.54 & 38.32 & 36.86 & 27.90 & 93.66 & 98.26 & 78.81 & 78.81 \\
  Perturb    & 97.50 & 98.86 & 99.95 & 99.10 & 64.01 & 40.45 & 41.14 & 31.93 & 87.01 & 91.46 & 98.66 & 91.38 \\
  Mixing     & 99.02 & 98.46 & 97.38 & 98.75 & 65.89 & 46.38 & 45.78 & 40.93 & 93.46 & 91.93 & 95.26 & 93.64 \\
  \bottomrule
  \end{tabular}

  \caption{Cross-source generalization on DetectRL, per-cell $F_1$
  (rows = training source, columns = evaluation source) across the
  Multi-Domain, Multi-LLM, and Multi-Attack settings.}
  \label{tab:train_test_detectrl_with_generalization_1_vector}
  \end{table*}

\subsection{MIRAGE}

Table~\ref{tab:test_mirage_3_vectors} reports the performance of SV-Detect and the baselines on the MIRAGE benchmark. Fig.~\ref{fig:mirage_bars} provides a visual summary of these results.

\begin{table*}[htbp]
\centering
\resizebox{\linewidth}{!}{
\begin{tabular}{l|cccc|cccc|cccc}
\toprule
\multicolumn{13}{c}{\textbf{MIRAGE-DIG (Disjoint-Input Generation)}} \\
\midrule
\multirow{2}{*}{Methods} & \multicolumn{4}{c|}{Generate} & \multicolumn{4}{c|}{Polish} & \multicolumn{4}{c}{Rewrite} \\
 & AUROC & Accuracy & MCC & TPR@5\% & AUROC & Accuracy & MCC & TPR@5\% & AUROC & Accuracy & MCC & TPR@5\% \\
\midrule
Likelihood        & 0.4936 & 0.5091 & 0.0183 & 0.0147 & 0.4653 & 0.5000 & 0.0000 & 0.0214 & 0.4337 & 0.5000 & 0.0000 & 0.0148 \\
LogRank           & 0.4992 & 0.5128 & 0.0260 & 0.0220 & 0.4512 & 0.5000 & 0.0000 & 0.0195 & 0.4225 & 0.5000 & 0.0000 & 0.0132 \\
Entropy           & 0.6522 & 0.6150 & 0.2543 & 0.1099 & 0.5543 & 0.5417 & 0.1247 & 0.0954 & 0.5805 & 0.5566 & 0.1650 & 0.1189 \\
RoBERTa-Base      & 0.5523 & 0.5397 & 0.1434 & 0.1250 & 0.4859 & 0.5010 & 0.0088 & 0.0460 & 0.5020 & 0.5049 & 0.0293 & 0.0569 \\
RoBERTa-Large     & 0.4716 & 0.5217 & 0.0842 & 0.0871 & 0.5171 & 0.5151 & 0.0340 & 0.0633 & 0.5570 & 0.5385 & 0.0864 & 0.0895 \\
LRR               & 0.5215 & 0.5341 & 0.0777 & 0.0701 & 0.4081 & 0.5000 & 0.0000 & 0.0200 & 0.3930 & 0.5000 & 0.0000 & 0.0188 \\
DNA-GPT           & 0.5733 & 0.5595 & 0.1196 & 0.0776 & 0.4771 & 0.5004 & 0.0110 & 0.0309 & 0.4453 & 0.5001 & 0.0080 & 0.0251 \\
NPR               & 0.6120 & 0.6140 & 0.2604 & 0.0191 & 0.5071 & 0.5370 & 0.1071 & 0.0318 & 0.4710 & 0.5201 & 0.0663 & 0.0226 \\
DetectGPT         & 0.6402 & 0.6258 & 0.2758 & 0.0275 & 0.5469 & 0.5531 & 0.1328 & 0.0355 & 0.5061 & 0.5266 & 0.0826 & 0.0283 \\
Fast-DetectGPT    & 0.7768 & 0.7234 & 0.4628 & 0.4310 & 0.5720 & 0.5570 & 0.1293 & 0.1189 & 0.5455 & 0.5432 & 0.1015 & 0.1025 \\
ImBD              & 0.8597 & 0.7738 & 0.5497 & 0.4065 & 0.7888 & 0.7148 & 0.4300 & 0.2730 & 0.7825 & 0.7068 & 0.4139 & 0.2933 \\
DetectAnyLLM            & 0.9525 & 0.8988 & 0.7975 & 0.7770 & 0.9297 & 0.8732 & 0.7487 & 0.7756 & 0.9234 & 0.8705 & 0.7447 & 0.7778 \\
SV-Detect (polish-500) & 0.9777 & 0.9241 & 0.8483 & 0.8862 & 0.9113 & 0.8364 & 0.6730 & 0.6414 & 0.9105 & 0.8332 & 0.6672 & 0.6437 \\
\textbf{SV-Detect (3-vec task-aware)} & \textbf{0.9912} & \textbf{0.9547} & \textbf{0.9095} & \textbf{0.9579} & \textbf{0.9509} & \textbf{0.8851} & \textbf{0.7706} & \textbf{0.7990} & \textbf{0.9503} & \textbf{0.8856} & \textbf{0.7713} & \textbf{0.7956} \\
\midrule
\multicolumn{13}{c}{\textbf{MIRAGE-SIG (Shared-Input Generation)}} \\
\midrule
\multirow{2}{*}{Methods} & \multicolumn{4}{c|}{Generate} & \multicolumn{4}{c|}{Polish} & \multicolumn{4}{c}{Rewrite} \\
 & AUROC & Accuracy & MCC & TPR@5\% & AUROC & Accuracy & MCC & TPR@5\% & AUROC & Accuracy & MCC & TPR@5\% \\
\midrule
Likelihood        & 0.4968 & 0.5207 & 0.0196 & 0.0145 & 0.4599 & 0.5002 & 0.0030 & 0.0233 & 0.4319 & 0.5000 & 0.0000 & 0.0111 \\
LogRank           & 0.5008 & 0.5183 & 0.0182 & 0.0186 & 0.4468 & 0.5000 & 0.0000 & 0.0211 & 0.4221 & 0.5000 & 0.0000 & 0.0118 \\
Entropy           & 0.6442 & 0.6123 & 0.1592 & 0.1074 & 0.5640 & 0.5439 & 0.0516 & 0.0946 & 0.5858 & 0.5645 & 0.0918 & 0.1198 \\
RoBERTa-Base      & 0.5368 & 0.5392 & 0.0529 & 0.1101 & 0.4741 & 0.5011 & 0.0048 & 0.0395 & 0.5099 & 0.5122 & 0.0221 & 0.0668 \\
RoBERTa-Large     & 0.4703 & 0.5236 & 0.0417 & 0.0910 & 0.5150 & 0.5157 & 0.0283 & 0.0702 & 0.5576 & 0.5426 & 0.0405 & 0.0762 \\
LRR               & 0.5214 & 0.5311 & 0.0314 & 0.0657 & 0.4076 & 0.5000 & 0.0000 & 0.0238 & 0.3978 & 0.5000 & 0.0000 & 0.0174 \\
DNA-GPT           & 0.5759 & 0.5647 & 0.0603 & 0.0813 & 0.4788 & 0.5001 & 0.0036 & 0.0340 & 0.4457 & 0.5002 & 0.0048 & 0.0258 \\
NPR               & 0.6088 & 0.6170 & 0.1571 & 0.0185 & 0.5074 & 0.5277 & 0.0612 & 0.0293 & 0.4738 & 0.5204 & 0.0340 & 0.0177 \\
DetectGPT         & 0.6353 & 0.6241 & 0.1719 & 0.0193 & 0.5434 & 0.5515 & 0.0668 & 0.0309 & 0.5079 & 0.5260 & 0.0431 & 0.0239 \\
Fast-DetectGPT    & 0.7706 & 0.7193 & 0.2078 & 0.4200 & 0.5727 & 0.5619 & 0.0607 & 0.1238 & 0.5480 & 0.5495 & 0.0525 & 0.1097 \\
ImBD              & 0.8612 & 0.7791 & 0.5599 & 0.4183 & 0.7951 & 0.7199 & 0.4451 & 0.3036 & 0.7694 & 0.6920 & 0.3936 & 0.2868 \\
DetectAnyLLM            & 0.9526 & 0.9059 & 0.8119 & 0.7722 & 0.9316 & 0.8740 & 0.7483 & 0.7779 & 0.9158 & 0.8643 & 0.7320 & 0.7574 \\
SV-Detect (polish-500) & 0.9779 & 0.9241 & 0.8488 & 0.8839 & 0.9089 & 0.8290 & 0.6583 & 0.6291 & 0.9039 & 0.8265 & 0.6532 & 0.6169 \\
\textbf{SV-Detect (3-vec task-aware)} & \textbf{0.9909} & \textbf{0.9516} & \textbf{0.9032} & \textbf{0.9516} & \textbf{0.9499} & \textbf{0.8837} & \textbf{0.7703} & \textbf{0.7811} & \textbf{0.9421} & \textbf{0.8749} & \textbf{0.7507} & \textbf{0.7706} \\
\bottomrule
\end{tabular}
}
\caption{Results across three tasks (\textsc{Generate}, \textsc{Polish}, \textsc{Rewrite}) under two evaluation settings (\textsc{MIRAGE-DIG} and \textsc{MIRAGE-SIG}) on MIRAGE. Metrics reported are AUROC, Balanced Accuracy, MCC, and TPR@5\%.}
\label{tab:test_mirage_3_vectors}
\end{table*}




\subsection{Ablation}

Tables~\ref{tab:ablation_on_classifier} and~\ref{tab:ablation_on_steering_vector_construction_method} provide results of ablating the choice of the downstream classifier and steering vector construction method. Sec.~\ref{sec:ablation} provides a summary of these results.

\begin{table*}[!ht]
  \centering
  \renewcommand{\arraystretch}{1}
  \setlength{\tabcolsep}{1pt}
  \resizebox{\textwidth}{!}{
  \begin{tabular}{l|cccc|cccc|cccc}
  \toprule
  \textbf{Classifier $\rightarrow$} &
  \multicolumn{4}{c|}{Logistic regression (base method)} &
  \multicolumn{4}{c|}{KNN ($n\_neighbors=5$)} &
  \multicolumn{4}{c}{CatBoost (default parameters)} \\
  \midrule
  \multicolumn{13}{c}{\textbf{Multi-Domain}} \\
  \midrule
  \textbf{Train $\downarrow$ / Eval $\rightarrow$} & ArXiv & XSum & Writing & Review & ArXiv & XSum & Writing & Review & ArXiv & XSum & Writing & Review \\
  \midrule
  ArXiv   & 100.00 &  97.34 &  86.82 &  94.05 & 100.00 &  50.25 &  50.00 &  50.50 & 100.00 &  96.61 &  92.39 &  96.65 \\
  XSum    &  97.80 &  99.94 &  83.02 &  95.24 &  87.13 &  99.65 &  72.37 &  84.70 &  97.52 &  99.93 &  83.21 &  94.42 \\
  Writing &  86.74 &  81.82 &  99.98 &  99.40 &  77.89 &  63.40 &  99.85 &  97.97 &  85.53 &  78.78 &  99.99 &  99.35 \\
  Review  &  94.17 &  93.96 &  99.38 &  99.96 &  53.90 &  64.96 &  80.85 &  99.74 &  93.85 &  91.24 &  99.10 &  99.96 \\
  \bottomrule
  \end{tabular}
  }
  \caption{Ablation on the downstream classifier on DetectRL Multi-Domain,
  per-cell AUROC (rows = training subset, columns = evaluation subset).}
  \label{tab:ablation_on_classifier}
  \end{table*}

\begin{table*}[!ht]
  \centering
  \renewcommand{\arraystretch}{1}
  \setlength{\tabcolsep}{1pt}
  \resizebox{\textwidth}{!}{
  \begin{tabular}{l|cccc|cccc|cccc}
  \toprule
  \textbf{Steering vector construction $\rightarrow$} &
  \multicolumn{4}{c|}{Logistic regression (base method)} &
  \multicolumn{4}{c|}{Mean difference} &
  \multicolumn{4}{c}{PCA} \\
  \midrule
  \multicolumn{13}{c}{\textbf{Multi-Domain}} \\
  \midrule
  \textbf{Train $\downarrow$ / Eval $\rightarrow$} & ArXiv & XSum & Writing & Review & ArXiv & XSum & Writing & Review & ArXiv & XSum & Writing & Review \\
  \midrule
  ArXiv   & 100.00 &  97.34 &  86.82 &  94.05 & 100.00 &  77.44 &  71.93 &  79.29 &  91.53 &  83.37 &  84.32 &  83.56 \\
  XSum    &  97.80 &  99.94 &  83.02 &  95.24 &  91.02 &  99.54 &  79.25 &  88.26 &  69.20 &  96.20 &  56.26 &  50.83 \\
  Writing &  86.74 &  81.82 &  99.98 &  99.40 &  85.27 &  90.26 &  99.06 &  97.16 &  76.18 &  82.37 &  93.82 &  92.69 \\
  Review  &  94.17 &  93.96 &  99.38 &  99.96 &  89.63 &  87.84 &  98.14 &  99.40 &  83.72 &  84.66 &  89.67 &  96.24 \\
  \bottomrule
  \end{tabular}
  }
  \caption{Ablation on the steering vector construction method on DetectRL
  Multi-Domain, per-cell AUROC (rows = training subset, columns = evaluation
  subset).}
  \label{tab:ablation_on_steering_vector_construction_method}
  \end{table*}

\subsection{Cross-benchmark transfer: bidirectional MIRAGE \texorpdfstring{$\leftrightarrow$}{<->} DetectRL}
\label{sec:appendix_cross_benchmark}

This subsection provides comprehensive results used for Fig.~\ref{fig:cross_benchmark_matrix} in the cross-benchmark transfer subsection in Sec.~\ref{sec:experiments}, broken down by all constituent settings and cells of each target benchmark. We compare SV-Detect against three zero-shot baselines (Fast-DetectGPT, Binoculars, RADAR) that use no training data on either benchmark, and two matched-supervision RoBERTa baselines fine-tuned on the same data as SV-Detect.

On the MIRAGE~$\to$~DetectRL side (Table~\ref{tab:bidirectional_MD_full_appendix}) we additionally report two multi-direction extensions of SV-Detect. The default variant \emph{polish-500} follows the setup in Sec.~\ref{sec:experiments_setup} and is trained on the original 500 human/machine pairs, where machine-generated side is GPT-3.5-Turbo-polished text. The
\emph{1-vec (union-800)} extends the training pool to 800 pairs, adding two task-specific subsets from~\citep{DBLP:journals/corr/abs-2412-10432}: 150 \textsc{generate} pairs, where the machine side is directly generated GPT-3.5-Turbo XSum text, and 150 \textsc{rewrite} pairs, where the machine side is a GPT-3.5-Turbo rewrite of the corresponding human text. \emph{1-vec (union-800)} keeps the
single-direction-per-layer setting, fitting one steering vector per layer on all 800 training pairs. The \emph{3-vec (task-aware)} setting fits a separate steering direction per training subset (\textsc{polish} / \textsc{generate} / \textsc{rewrite}) at each layer and combines them into an
orthonormal basis via QR decomposition. The matched-supervision RoBERTa baselines are fine-tuned on the same 500-pair default training set as the default \emph{polish-500} variant. Since Fast-DetectGPT, Binoculars, and RADAR
are zero-shot on DetectRL, their transfer numbers coincide with their in-domain scores on the DetectRL test split.

On the DetectRL~$\to$~MIRAGE side
(Table~\ref{tab:detectrl_D_M_full_appendix}) SV-Detect and both RoBERTa baselines are trained on the same joint union of the four DetectRL Multi-Domain training subsets. The task-aware \emph{4-vec} extension learns one steering direction per DetectRL Multi-Domain training subset and, following the extension protocol of MIRAGE, combines them into an orthonormal basis via QR decomposition.

\begin{table*}[t]
\centering
\scriptsize
\setlength{\tabcolsep}{3pt}
\renewcommand{\arraystretch}{1.03}
\resizebox{\textwidth}{!}{
\begin{tabular}{l | c c c | c c | c c c}
\toprule
& \multicolumn{3}{c|}{\textbf{Zero-shot}}
& \multicolumn{2}{c|}{\textbf{RoBERTa (matched: MIRAGE polish-500)}}
& \multicolumn{3}{c}{\textbf{SV-Detect variants (matched: MIRAGE)}} \\
\textbf{DetectRL test config}
    & Fast-DetectGPT & Binoculars & RADAR
    & RoBERTa-Base & RoBERTa-Large
    & polish-500 (default) & 1-vec (union-800) & 3-vec (task-aware) \\
\midrule
\multicolumn{9}{@{}l}{\emph{Multi-Domain}} \\
\hspace{1em}ArXiv               & 0.437 & 0.840 & 0.950 & 0.905 & 0.873 & 0.972 & 0.988 & 0.983 \\
\hspace{1em}XSum                & 0.392 & 0.774 & 0.998 & 0.891 & 0.905 & 0.960 & 0.964 & 0.963 \\
\hspace{1em}Writing             & 0.742 & 0.944 & 0.793 & 0.798 & 0.857 & 0.919 & 0.930 & 0.912 \\
\hspace{1em}Review              & 0.770 & 0.900 & 0.896 & 0.774 & 0.835 & 0.918 & 0.879 & 0.925 \\
\hspace{1em}\emph{Mean}         & \emph{0.585} & \emph{0.865} & \emph{0.909} & \emph{0.842} & \emph{0.867} & \emph{0.942} & \emph{0.940} & \emph{0.946} \\
\midrule
\multicolumn{9}{@{}l}{\emph{Multi-LLM}} \\
\hspace{1em}GPT-3.5             & 0.656 & 0.881 & 0.946 & 0.977 & 0.984 & 0.991 & 0.990 & 0.993 \\
\hspace{1em}Claude              & 0.300 & 0.552 & 0.811 & 0.692 & 0.693 & 0.884 & 0.900 & 0.867 \\
\hspace{1em}PaLM-2              & 0.660 & 0.933 & 0.951 & 0.786 & 0.813 & 0.902 & 0.918 & 0.874 \\
\hspace{1em}Llama-2             & 0.768 & 0.966 & 0.968 & 0.919 & 0.951 & 0.976 & 0.968 & 0.974 \\
\hspace{1em}\emph{Mean}         & \emph{0.596} & \emph{0.833} & \emph{0.919} & \emph{0.844} & \emph{0.860} & \emph{0.938} & \emph{0.944} & \emph{0.927} \\
\midrule
\multicolumn{9}{@{}l}{\emph{Multi-Attack} (mean over sub-attacks)} \\
\hspace{1em}Direct              & 0.796 & 0.949 & 0.922 & 0.907 & 0.912 & 0.973 & 0.973 & 0.970 \\
\hspace{1em}Prompt              & 0.784 & 0.935 & 0.887 & 0.902 & 0.907 & 0.960 & 0.959 & 0.957 \\
\hspace{1em}Paraphrase          & 0.701 & 0.883 & 0.943 & 0.818 & 0.833 & 0.939 & 0.944 & 0.940 \\
\hspace{1em}Perturbation        & 0.496 & 0.769 & 0.922 & 0.832 & 0.856 & 0.923 & 0.934 & 0.909 \\
\hspace{1em}Mixing              & 0.672 & 0.891 & 0.935 & 0.885 & 0.875 & 0.952 & 0.956 & 0.944 \\
\hspace{1em}\emph{Mean}         & \emph{0.690} & \emph{0.885} & \emph{0.922} & \emph{0.869} & \emph{0.876} & \emph{0.949} & \emph{0.953} & \emph{0.944} \\
\midrule
\textbf{Grand mean (13 configs)}
    & \textbf{0.629} & \textbf{0.863} & \textbf{0.917}
    & \textbf{0.852} & \textbf{0.866}
    & \textbf{0.943} & \textbf{0.947} & \textbf{0.940} \\
\bottomrule
\end{tabular}
}
\caption{MIRAGE~$\to$~DetectRL cross-benchmark transfer, per-cell AUROC.
Multi-Attack rows are unweighted means over the DetectRL sub-attack
categories (Direct: 1 test set; Prompt: 3 sub-attacks; Paraphrase / Perturbation:
4 each; Mixing: 3).}
\label{tab:bidirectional_MD_full_appendix}
\end{table*}

\begin{table*}[t]
\centering
\scriptsize
\setlength{\tabcolsep}{4pt}
\renewcommand{\arraystretch}{1.05}
\begin{tabular}{l | c c c | c c | c c}
\toprule
& \multicolumn{3}{c|}{\textbf{Zero-shot}}
& \multicolumn{2}{c|}{\textbf{RoBERTa (matched: DetectRL joint)}}
& \multicolumn{2}{c}{\textbf{SV-Detect variants (matched: DetectRL)}} \\
\textbf{MIRAGE test setting}
    & Fast-DetectGPT & Binoculars & RADAR
    & RoBERTa-Base & RoBERTa-Large
    & 1-vec (default) & 4-vec (task-aware) \\
\midrule
DIG-Generate & 0.777 & 0.890 & 0.865 & 0.928 & 0.964 & \textbf{0.975} & 0.952 \\
DIG-Polish   & 0.572 & 0.624 & 0.657 & 0.844 & \textbf{0.897} & 0.859 & 0.831 \\
DIG-Rewrite  & 0.546 & 0.608 & 0.685 & 0.839 & \textbf{0.894} & 0.852 & 0.821 \\
SIG-Generate & 0.771 & 0.890 & 0.868 & 0.924 & 0.964 & \textbf{0.975} & 0.946 \\
SIG-Polish   & 0.573 & 0.620 & 0.649 & 0.848 & \textbf{0.898} & 0.857 & 0.834 \\
SIG-Rewrite  & 0.548 & 0.604 & 0.681 & 0.836 & \textbf{0.890} & 0.846 & 0.816 \\
\midrule
\textbf{Mean}
    & \textbf{0.631} & \textbf{0.706} & \textbf{0.734}
    & \textbf{0.870} & \textbf{0.918}
    & \textbf{0.894} & \textbf{0.867} \\
\bottomrule
\end{tabular}
\caption{DetectRL~$\to$~MIRAGE reverse transfer, per-cell AUROC across the
six MIRAGE test settings. Bold marks the winning method per row.}
\label{tab:detectrl_D_M_full_appendix}
\end{table*}

\subsection{Cross-benchmark transfer to PADBen and RAID}
\label{sec:appendix_zeroshot_padben}

Table~\ref{tab:zeroshot_appendix} reports per-task AUROC on PADBen's five sentence-pair regimes alongside the RAID TPR@FPR=5\% Chat / Non-Chat / Overall summary. Both benchmarks are used purely as \emph{targets}, and no training data from either is seen by any of the detectors.

The two \emph{default} rows correspond to the default SV-Detect setup described in Sec.~\ref{sec:experiments_setup}: on MIRAGE, steering vectors are trained on the 500 human/machine pairs, where machine-generated side is GPT-3.5-Turbo-polished text (\emph{MIRAGE polish-500}), and on DetectRL, steering vectors are trained on the joint union of the four Multi-Domain training subsets(\emph{DetectRL 1-vec}). As in Sec.~\ref{sec:appendix_cross_benchmark}, we additionally report three multi-direction extensions: \emph{MIRAGE 1-vec (union-800)} extends the MIRAGE training pool to 800 pairs as described in Sec.~\ref{sec:appendix_cross_benchmark}, while keeping a single steering direction per
layer; \emph{MIRAGE 3-vec (task-aware)} fits one steering direction per extended MIRAGE training data (Generate / Polish / Rewrite) and combines them via
QR decomposition; \emph{DetectRL 4-vec (task-aware)} does the same on the DetectRL side with one direction per Multi-Domain training subset.

On PADBen, our detectors outperform every reported zero-shot baseline. Note that the IPD task is near-chance for every method including all four PADBen paper
baselines, suggesting that this task is universally hard for all the detectors. On RAID, our default
single-direction detectors are competitive on chat-tuned generators but weak on base-model (Non-Chat) generators --- consistent with our training data containing only chat-tuned model outputs. The task-aware multi-direction extensions substantially close this gap: DetectRL~4-vec
reaches $56.3\%$ Overall TPR@$5\%$ and $78.9\%$ on Chat generators.

\begin{table*}[t]
\centering
\small
\setlength{\tabcolsep}{4pt}
\renewcommand{\arraystretch}{1.05}
\begin{tabular}{l | c c c | c c c c c c}
\toprule
& \multicolumn{3}{c|}{\textbf{RAID TPR@FPR=5\%}}
& \multicolumn{6}{c}{\textbf{PADBen sentence-pair AUROC per task}} \\
\textbf{Detector}
    & Chat & Non-Chat & Overall
    & PSA & GAD & AIT-L & IPD & DPA & Mean \\
\midrule
\multicolumn{10}{@{}l}{\emph{Zero-shot baselines on the target benchmark}} \\
LLMDet          & 37.4 & 27.6 & 32.5 & --    & --    & --    & --    & --    & --    \\
GLTR            & 69.9 & 50.4 & 60.1 & 0.429 & 0.436 & 0.401 & 0.514 & 0.488 & 0.454 \\
Fast-DetectGPT  & 77.5 & 58.4 & 67.9 & 0.638 & 0.787 & 0.504 & 0.478 & 0.369 & 0.555 \\
RADAR           & 79.5 & 58.9 & 69.2 & 0.728 & 0.910 & 0.748 & 0.526 & 0.909 & 0.764 \\
Binoculars (Falcon-7B) & 86.7 & 58.8 & 72.8 & 0.399 & 0.457 & 0.505 & 0.498 & 0.457 & 0.463 \\
\midrule
\multicolumn{10}{@{}l}{\emph{Ours (default single-direction)}} \\
SV-Detect (MIRAGE polish-500) & 71.4 & 10.2 & 40.8 & 0.890 & 0.943 & 0.778 & 0.498 & 0.949 & 0.812 \\
SV-Detect (DetectRL 1-vec)    & 65.4 & 26.7 & 46.1 & 0.944 & 0.968 & 0.790 & 0.524 & 0.978 & 0.841 \\
\midrule
\multicolumn{10}{@{}l}{\emph{Ours (multi-direction extensions)}} \\
SV-Detect (MIRAGE 1-vec union-800)   & 71.7 &  9.4 & 40.5 & 0.886 & 0.942 & 0.753 & 0.494 & 0.939 & 0.803 \\
SV-Detect (MIRAGE 3-vec task-aware)  & 73.9 & 13.2 & 43.6 & 0.931 & 0.952 & 0.762 & 0.503 & 0.969 & 0.823 \\
SV-Detect (DetectRL 4-vec task-aware)& 78.9 & 33.7 & 56.3 & 0.876 & 0.948 & 0.782 & 0.509 & 0.950 & 0.813 \\
\bottomrule
\end{tabular}
\caption{Full zero-shot cross-benchmark results.
\textbf{RAID:} TPR@FPR=5\%, macro-averaged over the 8 open-source
configurations, split into Chat (llama-chat, mistral-chat, mpt-chat) and
Non-Chat (gpt2, mistral, mpt) LLM families. \textbf{PADBen:} sentence-pair AUROC per task (PSA = paraphrase-source attribution, GAD = general authorship detection, AIT-L = AI-text laundering, IPD = iterative paraphrase depth, DPA = deep-paraphrase attack)}
\label{tab:zeroshot_appendix}
\end{table*}

\clearpage
\section{Additional comparison with RepreGuard}
\label{sec:appendix_repreguard_vs_svdetect}
RepreGuard~\citep{RepreGuard} is the closest representation-based AI-generated text detection method to SV-Detect. 

Table~\ref{tab:train_test_detectrl_without_generalization_1_vector} reports results of RepreGuard under our DetectRL evaluation protocol using the GPT-Neo-2.7B backbone.
However, the default backbone of RepreGuard is LLaMA-3.1-8B.
In this section, we provide an additional comparison of SV-Detect to RepreGuard under the matched backbones using the default setup of ~\citet{RepreGuard}: 512-pair train/test cells with per-LLM stratification, and macro-averaging across a small number of bootstrap seeds. We use RAID~\citep{raid} for comparison, which is the default benchmark used in ~\citet{RepreGuard}.

Table~\ref{tab:raid_main} reports RAID performance for RepreGuard and SV-Detect trained on RAID official training split, under RepreGuard's training protocol, at both GPT-Neo-2.7B and LLaMA-3.1-8B-Instruct. We follow the aggregation of \citet{raid} and macro-average
each row across the eight open-source RAID configurations (Chat vs.\ Non-Chat $\times$ greedy vs.\ sampling $\times$ repetition-penalty on/off).

Results suggest that RepreGuard is highly backbone-sensitive: at GPT-Neo-2.7B its RAID score collapses to near-chance (Chat $4.5\%$ TPR@$5\%$, Non-Chat $5.3\%$; AUROC $\approx\!0.5$), and only moving to LLaMA-3.1-8B-Instruct recovers Chat performance ($92.9\%$ TPR@$5\%$). Non-Chat generators remain hard for RepreGuard even at LLaMA-8B ($41.7\%$). The same backbone-sensitivity is visible on DetectRL Multi-LLM: our reproduction of RepreGuard under its own protocol yields $2.5\%$ TPR@$5\%$ (AUROC $0.366$) at GPT-Neo-2.7B and $30.3\%$ TPR@$5\%$ (AUROC $0.812$) at
LLaMA-3.1-8B-Instruct, macro-averaged across the four Multi-LLM generators. SV-Detect does not exhibit this dependence: at both GPT-Neo-2.7B and LLaMA-3.1-8B it reaches $\approx\!99.7\%$ TPR@$5\%$ / $\approx\!1.0$ AUROC on both Chat and Non-Chat RAID cells. In other words, SV-Detect matches or beats RepreGuard at its \emph{best} backbone while remaining strong at the small GPT-Neo backbone where RepreGuard collapses. This supports our claim that the LogReg-based steering construction produces a
discriminative direction that is robust to backbone choice, whereas RepreGuard's cluster-separation heuristic degrades sharply when the observer LLM's representations are less well-separated.

%
%
%
%
\begin{table}[t]
\centering
\small
\setlength{\tabcolsep}{5pt}
\renewcommand{\arraystretch}{1.05}
\begin{tabular}{l l c c}
\toprule
\textbf{Method} & \textbf{Backbone} & \textbf{TPR@5\%} & \textbf{AUROC} \\
\midrule
\multicolumn{4}{@{}l}{\emph{Zero-shot / score-based (RAID Table~5)}} \\
LLMDet          & --            & 32.5  & --     \\
GLTR            & --            & 60.1  & --     \\
Fast-DetectGPT  & --            & 67.9  & --     \\
RADAR           & --            & 69.2  & --     \\
Binoculars      & Falcon-7B     & 72.8  & --     \\
\midrule
\multicolumn{4}{@{}l}{\emph{Supervised, applied zero-shot to RAID (RAID Table~5)}} \\
RoBERTa-Base (GPT-2)  & --      & 59.9  & --     \\
\midrule
\multicolumn{4}{@{}l}{\emph{Trained on RAID}} \\
RepreGuard      & GPT-Neo-2.7B  & 4.9   & 0.5500 \\
RepreGuard      & LLaMA-3.1-8B  & 93.2  & 0.9634 \\
\textbf{SV-Detect (ours)} & GPT-Neo-2.7B & \textbf{99.7} & \textbf{0.9993} \\
\textbf{SV-Detect (ours)} & LLaMA-3.1-8B & \textbf{99.7} & \textbf{0.9993} \\
\bottomrule
\end{tabular}
\caption{RAID performance on the eight open-source configurations (Chat / Non-Chat $\times$ decoding $\times$ repetition penalty).}
\label{tab:raid_main}
\end{table}

\end{document}